\documentclass[runningheads]{llncs}

\usepackage{eccv}

\usepackage{graphicx}
\usepackage{amsmath}
\usepackage{amssymb}
\usepackage{booktabs}
\usepackage{enumitem}
\usepackage{appendix}
\usepackage[percent]{overpic}
\usepackage{tikz}
\usepackage{multicol} 
 \usepackage{vwcol}  
 \usepackage{paracol}

\usepackage[pagebackref,breaklinks,colorlinks]{hyperref}

\usepackage[capitalize]{cleveref}
\crefname{section}{Sec.}{Secs.}
\Crefname{section}{Section}{Sections}
\Crefname{table}{Table}{Tables}
\crefname{table}{Tab.}{Tabs.}

\usepackage{eccvabbrv}
\usepackage[accsupp]{axessibility}
\usepackage{wrapfig}

\newcommand{\methodname}{$\phi$-PD}
\usepackage{animate}
\usepackage{graphicx, animate}

\makeatletter
  \def\title@font{\Large\bfseries}
  \let\ltx@maketitle\@maketitle
  \def\@maketitle{\bgroup%
    \let\ltx@title\@title%
    \def\@title{\resizebox{\textwidth}{!}{%
      \mbox{\title@font\ltx@title}%
    }}%
    \ltx@maketitle%
  \egroup}
\makeatother

\begin{document}

\title{NeuralRemaster: Phase-Preserving Diffusion for Structure-Aligned Generation}
\author{Yu Zeng\inst{1} \and Charles Ochoa\inst{1} \and  Mingyuan Zhou\inst{2} \and  Vishal M. Patel\inst{3} \and \\ Vitor Guizilini\inst{1} \and  Rowan McAllister\inst{1}
}
\institute{Toyota Research Institute \and University of Texas, Austin \and Johns Hopkins University}


\titlerunning{Phase-Preserving Diffusion} 
\authorrunning{Y.~Zeng et al.}

\maketitle
\begin{center}
    \centering
    \includegraphics[width=\textwidth]{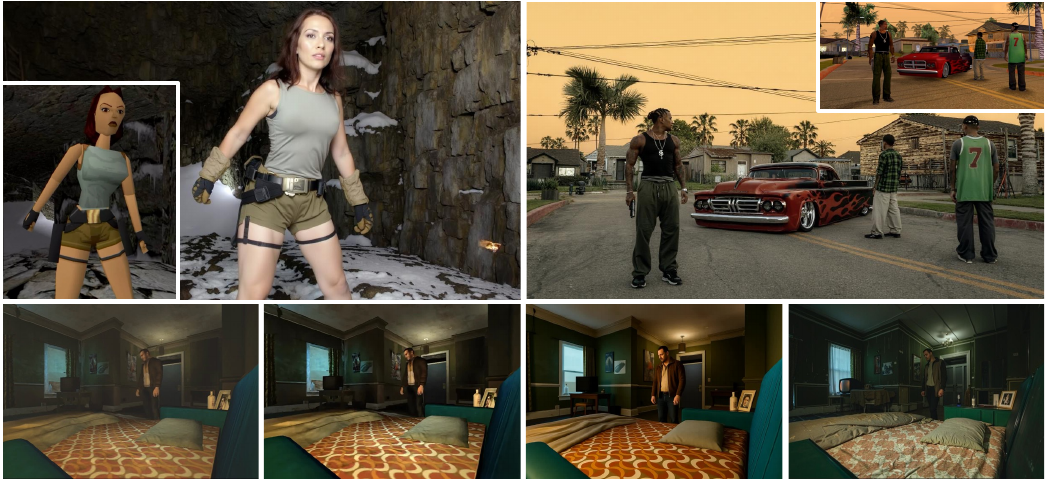} \\
    \vspace{-5pt}
    {\scriptsize\hfill{Input\textcolor{white}{~ e~~~~~~~~~~~~~}} \hfill  {FLUX-Kontext} \hfill {\textcolor{white}{~~~~~}QWen-Edit\textcolor{white}{~~}} \hfill {\textcolor{white}{~~~~~~~~~~~~~}Ours} \hfill ~}
\end{center}%
\vspace{-15pt}
\begin{center}
    \includegraphics[trim={0 1mm 0 0},clip,width=\textwidth]{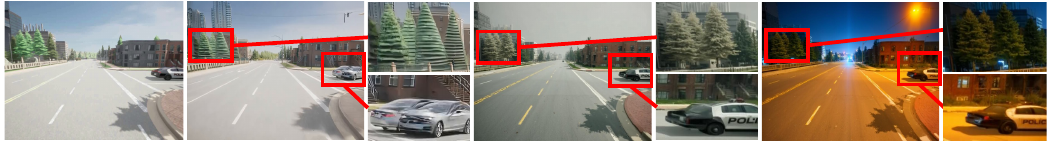}
\end{center}%
\vspace{-15pt}
    {\scriptsize 
    \hspace{6mm} Input
    \hspace{10mm} Cosmos-Transfer 2.5
    \hspace{14mm} Ours
    \hspace{22mm} Ours}
\vspace{-10pt}
\begin{center}
    \captionof{figure}{\small
    We present Phase-Preserving Diffusion (\methodname{}), a model-agnostic reformulation of the diffusion process that preserves an image's phase while randomizing its magnitude, enabling structure-aligned generation with \textbf{no architectural changes or additional parameters}.
    }
    \label{fig:teaser}
\end{center}%

\begin{abstract} 
Standard diffusion corrupts data using Gaussian noise whose Fourier coefficients have random magnitudes and random phases. 
While effective for unconditional or text-to-image generation, corrupting phase components destroys spatial structure, making it ill-suited for tasks requiring geometric consistency, such as re-rendering, simulation enhancement, and image-to-image translation. 
We introduce Phase-Preserving Diffusion (\methodname), a model-agnostic reformulation of the diffusion process that preserves input phase while randomizing magnitude, enabling structure-aligned generation without architectural changes or additional parameters. 
We further propose Frequency-Selective Structured (FSS) noise, which provides continuous control over structural rigidity via a single frequency-cutoff parameter.
\methodname{} adds no inference-time cost and is compatible with any diffusion model for images or videos. 
Across photorealistic and stylized re-rendering, as well as sim-to-real enhancement for driving planners, \methodname{} produces controllable, spatially aligned results. When applied to the CARLA simulator, \methodname{} significantly improves sim-to-real planner transfer performance. The method is complementary to existing conditioning approaches and broadly applicable to image-to-image and video-to-video generation. Videos, additional examples, and code are available on our \href{https://yuzeng-at-tri.github.io/ppd-page/}{project page}.
\end{abstract}

\section{Introduction}
\label{sec:intro}
Recent advances in diffusion models have revolutionized image generation, achieving high-fidelity results for unconditional or text-conditioned synthesis. 
Yet many practical applications do not require generating a scene from scratch. Instead, they operate within an image-to-image setting where the spatial layout, such as object boundaries, geometry and scene structures, should remain fixed while the appearance is modified. 
Examples include neural rendering, stylization, and sim-to-real transfer for autonomous driving or robotics simulation. We refer to this broad class of problems as \textbf{structure-aligned generation}. 

Although these tasks are conceptually easier than generating from scratch, existing solutions are unnecessarily complex. 
Methods such as ControlNet \cite{zhang2023adding}, T2I-Adapter \cite{mou2024t2i}, and related variants attach auxiliary branches to inject structural input into the model. While effective, this introduces additional parameters and computational cost, paradoxically making structure-aligned generation harder than it should be. 

We argue that this inefficiency stems not from the network architecture, but from the diffusion process itself.
The forward diffusion process injects Gaussian noise, which destroys both the magnitude and phase components in the frequency domain.  
Classical signal processing \cite{oppenheim1981importance,ruderman1994statistics,wang2005translation}, however, tells us that phase encodes structure while magnitude encodes texture. 
Destroying the phase means destroying the very spatial coherence that structure-aligned generation depends on, forcing the model to reconstruct structure from scratch.

\begin{figure*}[t!]
  \centering
  \includegraphics[width=\linewidth]{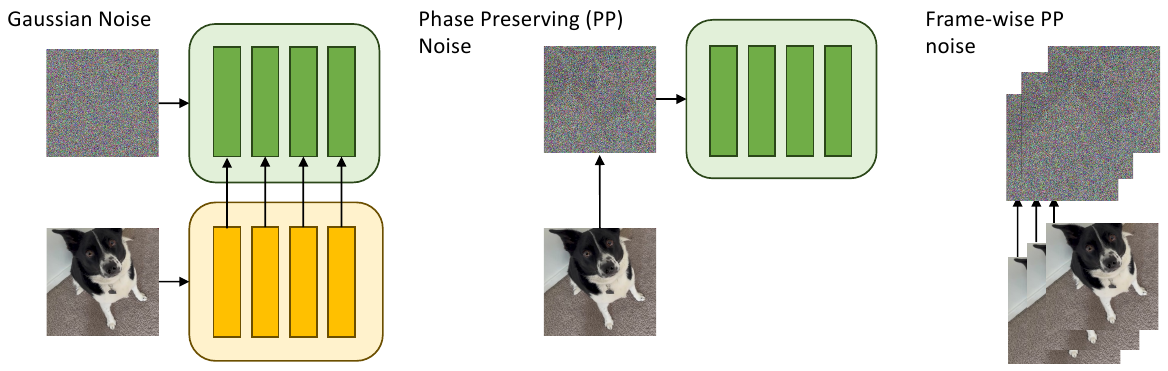}
  {\scriptsize
  \noindent 
  \begin{minipage}[t]{0.35\textwidth} 
    Prior methods encode structural input with additional modules (yellow), that depend on the model (green) and incur additional computation.
  \end{minipage}%
  \hfill 
  \begin{minipage}[t]{0.3\textwidth} 
    Our method incurs no additional overhead and works with any model (green).
  \end{minipage}
  \hfill 
  \begin{minipage}[t]{0.25\textwidth} 
    Our method extends to video.
  \end{minipage}
  }
   \caption{Unlike prior approaches that modify architectures and add overhead, \methodname{} preserves structure via phase consistency, remaining lightweight and model-agnostic, reflecting that image-conditioned generation should be simpler, not harder. }
   \label{fig:main-method}
\end{figure*}

Motivated by this insight, we propose Phase-Preserving Diffusion (\methodname). Instead of corrupting data with Gaussian noise, \methodname{} constructs \textit{structured noise} whose magnitude matches that of Gaussian noise while preserving the input phase. This naturally maintains spatial alignment throughout sampling (Figure~\ref{fig:teaser}) with \textbf{no architectural modification, no extra parameters} (Figure~\ref{fig:main-method}), and is compatible with any DDPM or flow-matching model for images or videos. 

To provide controllable levels of structural rigidity, we further introduce \textbf{Frequency-Selective Structured (FSS) noise}, which interpolates between input phase and pure Gaussian noise via a single cutoff parameter (Figure \ref{fig:noise-visualize}). This allows us to control the trade-off between strict alignment and creative flexibility. 

We evaluate \methodname{} across photorealistic re-rendering, stylized re-rendering and simulation enhancement for embodied-AI agents. \methodname{} consistently maintains geometry alignment while producing high-quality visual outputs, outperforming prior methods across both quantitative and qualitative metrics. 
When used to enhance CARLA simulations, \methodname{} improves planner transfer to the Waymo Open Dataset by 50\%, substantially narrowing the sim-to-real gap.
In summary, our contributions include:
\begin{itemize}[noitemsep]
    \item \textbf{Phase-preserving diffusion process:}  A diffusion process that preserves phase while randomizing magnitude in frequency domain, maintaining spatial structure without architectural changes.
    \item \textbf{Frequency-selective structured noise:} A single-parameter mechanism that enables continuous control over structural alignment rigidity.
    \item \textbf{Unified and efficient framework} applicable to both images and videos, compatible with DDPMs and flow-matching, and requires no inference-time overhead. 
\end{itemize}

\section{Related Work}
\noindent \textbf{Diffusion Models. } 
Diffusion models have become a dominant paradigm for generative modeling, capable of representing complex data distributions with remarkable fidelity~\cite{ho2020denoising, karras2022elucidating}. 
They progressively corrupt data into Gaussian noise through a forward diffusion process, then learn to invert this process via iterative denoising. 
This framework has demonstrated state-of-the-art performance across diverse domains, including image, video, and audio generation~\cite{saharia2022photorealistic, ramesh2022hierarchical, balaji2022ediff, chen2023videocrafter1, ho2022video, popov2021grad, kong2020diffwave}, as well as reinforcement learning~\cite{janner2022planning, wang2022diffusion, psenka2023learning} and robotics~\cite{ajay2022conditional, urain2023se, chi2023diffusion}.

\noindent \textbf{Frequency-Domain Manipulation for Diffusion. }
Recent work has explored frequency domain operations for diffusion models. \cite{dielemanblog} argues that diffusion models of images perform approximate autoregression in the frequency domain. 
\cite{falck2025fourier} shows that the standard diffusion forward process corrupts high-frequency components faster than low-frequency ones, and introduces an alternate process that corrupts all frequencies at the same rate. 
\cite{yu2025dmfft} shows that modifying the UNet frequency domain features significantly improves the generating quality for image or video generation. 
FreeDiff \cite{wu2024freediff} introduces a fine-tuning free approach for image editing that employs progressive frequency truncation to refine the guidance of diffusion models. 
\cite{zhang2025training} proposes a training-free style transfer method that modulates the intermediate samples with the image phase.  
\cite{qian2024boosting} proposed to use a frequency-dependent moving average during sampling. 
\cite{baek2025sonic} proposes a training-free approach for image inpainting that optimizes the initial seed noise in the spectral domain. 

\noindent \textbf{Structure-Aligned Generation with Diffusion.} 
Most existing methods achieve structure-aligned generation by modifying the network architecture and introducing additional adaptation components. 
ControlNet \cite{zhang2023adding} copies the entire U-Net encoder into a trainable encoder branch, which adds significant computation overhead. T2I-Adapter \cite{mou2024t2i} reduces computation overhead using a lightweight adapter module but sacrifices control precision. Uni-ControlNet \cite{zhao2023unicontrolnet} enables simultaneous utilization of multiple local controls by training two adapters. OmniControl \cite{tan2024ominicontrol2} integrates image conditions into Diffusion Transformer (DiT) architectures with only 0.1\% additional parameters by re-using the VAE and transformer blocks of the base model. ControlNeXt \cite{peng2024controlnext} uses a lightweight convolutional module to inject control signals, and directly finetune selective parameters of the base model to reduce training costs and latency increase. SCEdit \cite{jiang2024scedit} proposes an efficient finetuning framework that edits skip connections using a lightweight module. NanoControl \cite{huang2024nanocontrol} aims to achieve efficient control with a LoRA-style control module. The above methods all rely on an additional module to incorporate the control signal, though some are more lightweight than others.  CosmosTransfer \cite{Alhaija2025CosmosTransfer1CW} achieves multi-modal control by combining multiple ControlNet branches, demonstrating promising applications on physical tasks; however, multiple branches introduce significant computation overhead.
In contrast, \methodname{} does not introduce any computation overhead or additional parameters while enabling universal spatial control. 

\noindent \textbf{Training-Free Guidance Methods. } Recently, several training-free methods have been developed. \cite{yu2023freedom} introduced FreeDoM, which leverages off-the-shelf pre-trained networks to construct time-independent energy functions that guide generation. \cite{couairon2023zeroshot} proposed ZestGuide for zero-shot spatial layout conditioning, utilizing implicit segmentation maps extracted from cross-attention layers to align generation with input masks. \cite{mo2024freecontrol} presented FreeControl, a training-free approach that enforces structure guidance with the base model feature extracted from the control signal. Although these methods avoid training cost, they introduce additional overhead at test time, either an external model, DDIM inversion, or multiple inferences of the base model. In contrast, \methodname{} achieves spatial control without any additional inference time overhead. 

\section{Method}
\subsection{Frequency Domain Fundamentals} 
In the frequency domain, any image $I(x,y)$ can be represented through the 2D Fourier transform:
\begin{align}
    F(u,v) &= \mathcal{F}\{I(x,y)\} = \sum_{x=0}^{W-1}\sum_{y=0}^{H-1} I(x,y) e^{-2\pi j(ux/W + vy/H)},
\end{align}
where $F(u,v)$ is a complex-valued function that can be decomposed into magnitude and phase components:
\begin{equation}
    F(u,v) = |F(u,v)| \cdot e^{j\phi(u,v)} = A(u,v) \cdot e^{j\phi(u,v)}.
\end{equation}
Here, $A(u,v) = |F(u,v)|$ is the magnitude spectrum and $\phi(u,v)$ the phase spectrum.
The inverse Fourier transform uses magnitude and phase to reconstruct the original image:
\begin{align}
    I(x,y) &= \mathcal{F}^{-1}\{F(u,v)\} = \sum_{u=0}^{W-1}\sum_{v=0}^{H-1} F(u,v) e^{2\pi j (ux/W + vy/H)}.
\end{align}
\begin{wrapfigure}{r}{0.5\textwidth}
   \vspace{-20pt}
  \centering
   \includegraphics[width=\linewidth]{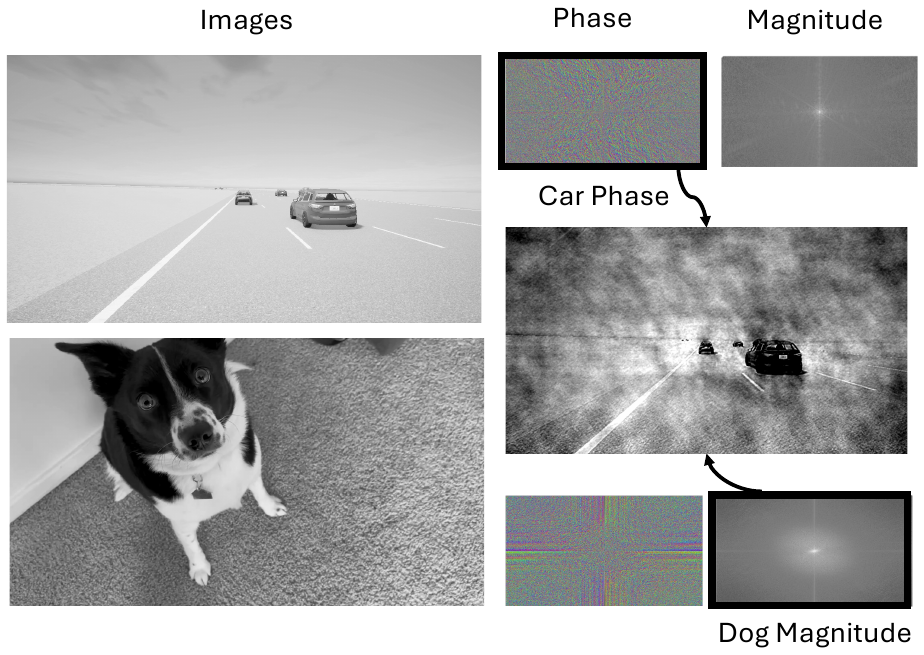}
   \caption{Mixing phase and magnitude from two images. The mixture keeps the structure of the image where the phase is taken.}
   \vspace{-20pt}
   \label{fig:phase-mix}
\end{wrapfigure}

\noindent \textbf{Phase-Magnitude Separation in Signal Processing.} 
Foundational work by Oppenheim \cite{oppenheim1981importance} shows that phase primarily determines spatial structure, while magnitude largely controls texture statistics. Mixing phase and magnitude from different images produces reconstructions whose spatial layout follows the source of the phase, not magnitude (see Figure~\ref{fig:phase-mix}). This observation motivates our approach: if diffusion destroys phase, it destroys spatial geometry; if we preserve phase, we preserve structure. 

\subsection{Phase-Preserving Diffusion}
Standard diffusion corrupts data using Gaussian noise whose Fourier coefficients have random magnitudes and random phases. As a result, even early diffusion steps erase spatial alignment. We propose a simple alternative: \emph{preserve the input image's phase and randomize the magnitude}, by using structured noise that shares the input phase. 

\noindent \textbf{Structured Noise Construction.}
Given an input image $I$, we compute its Fourier transform:
\begin{equation}
    F_I = A_I \cdot e^{j\phi_I}.
\end{equation}
We construct \textbf{phase-preserving noise} by pairing the input image phase with a random magnitude:
\begin{equation}
    F_{\hat{\epsilon}} = A_{\epsilon} \cdot e^{j\phi_I},
\end{equation}
and invert it:
\begin{equation}
    \hat{\epsilon} = \mathcal{F}^{-1}\{F_{\hat{\epsilon}}\},
\end{equation}
where the random magnitude $A_\epsilon$ can be obtained from the Fourier transform of Gaussian noise:
\begin{equation}
A_\epsilon = |\mathcal{F}\{\epsilon\}|, \quad \epsilon \sim \mathcal{N}(0, I).
\end{equation}
Alternatively, $A_\epsilon$ can be sampled from a scaled Rayleigh distribution~\cite{GoodmanStatisticalOptics}:
\begin{equation}
A_\epsilon = \sqrt{-N \ln U}, \quad U \sim \text{Uniform}(0, 1),
\end{equation}
where $N = H \times W$, matching the magnitude statistics of the DFT of unit-variance Gaussian noise.

This structured noise is used in place of Gaussian noise in forward diffusion for training. It injects randomness while maintaining the phase of the input. At test time, we achieve structure-aligned generation by starting sampling from structured noise constructed with input image phase. 

\noindent \textbf{Frequency Selective Structured (FSS) Noise.}
In practice, we often want to control to what extent we keep the structure from the input image. Some tasks require strict structure preservation, while others benefit from partial freedom to reinterpret the scene. To provide this control, we introduce Frequency Selective Structured (FSS) noise, which only keep the image within a radius $r$  and use the phase from the noise for the remainder. We define a smooth frequency mask $M(u,v)$ based on the cutoff radius $r$:
\begin{equation}
    M(u,v) = \begin{cases}
    1 & \text{if } \sqrt{u^2 + v^2} \leq r \\
    \exp\left(-\frac{(\sqrt{u^2 + v^2} - r)^2}{2\sigma^2}\right) & \text{if } \sqrt{u^2 + v^2} > r
    \end{cases}
\end{equation}
where $(u, v)$ are centered frequency coordinates with the DC component at the origin, and $r$ is specified in pixels relative to the frequency grid of size $W \times H$. In practice, we use \texttt{fftshift} to center the spectrum before applying the mask. 

The FSS noise $\hat{\epsilon}$ is the combination of image phase and noise phase using the mask:
\begin{equation}
    F_{\hat{\epsilon}} = A_{\epsilon} \cdot e^{j\phi_I\odot M +j\phi_\epsilon\odot(1-M)},
\end{equation}
where $\odot$ represents element-wise multiplication. 
We can sample phase $\phi_\epsilon$ as the Fourier transform of Gaussian noise:
\begin{align}
    \epsilon \;&\sim\; \mathcal{N}(0,1), \nonumber \\
    \phi_\epsilon \;&=\; \text{arg}(\mathcal{F}\{ \epsilon\}) \,\sim\, \text{Uniform}(-\pi,\pi).
\end{align}

\begin{figure}[t!]
  \centering
   \includegraphics[width=0.18\linewidth]{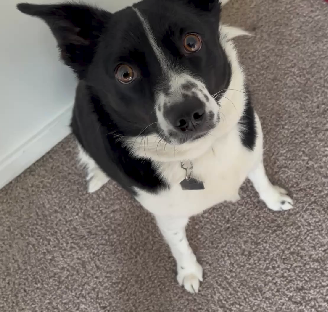} 
   \includegraphics[width=0.18\linewidth]{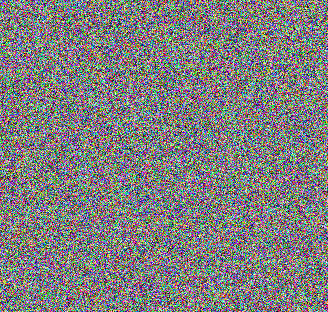} 
   \includegraphics[width=0.18\linewidth]{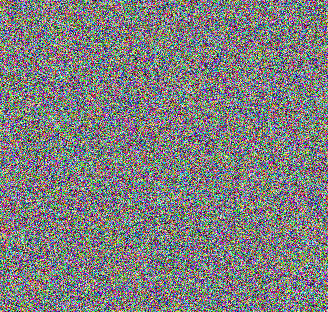} 
   \includegraphics[width=0.18\linewidth]{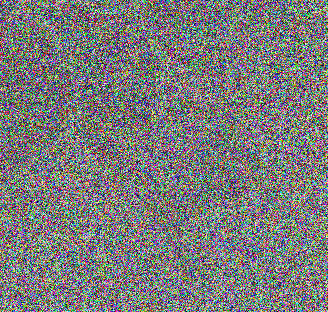} 
   \caption{Frequency Selective Structured (FSS) Noise with increasing cutoff radius $r$. }
   \label{fig:noise-visualize}
   \vspace{10pt}
   \includegraphics[width=0.16\linewidth,height=0.16\linewidth]{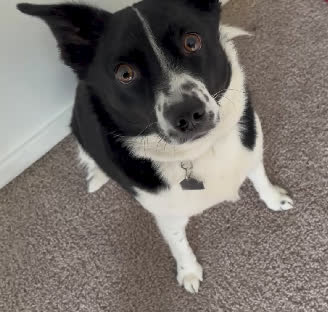} 
   \includegraphics[width=0.16\linewidth]{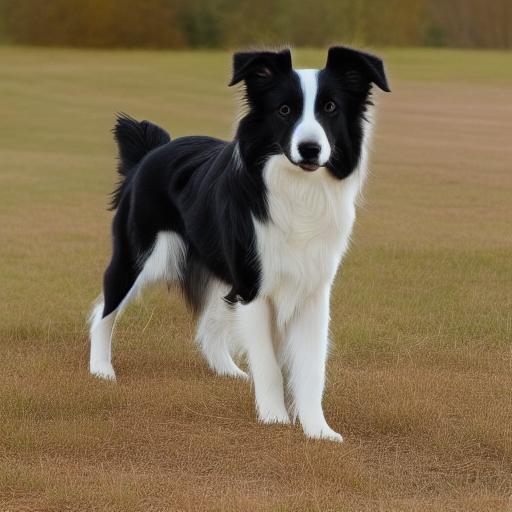} 
   \includegraphics[width=0.16\linewidth]{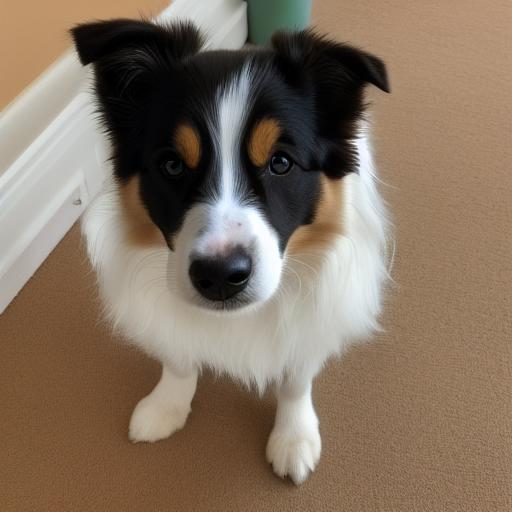} 
   \includegraphics[width=0.16\linewidth]{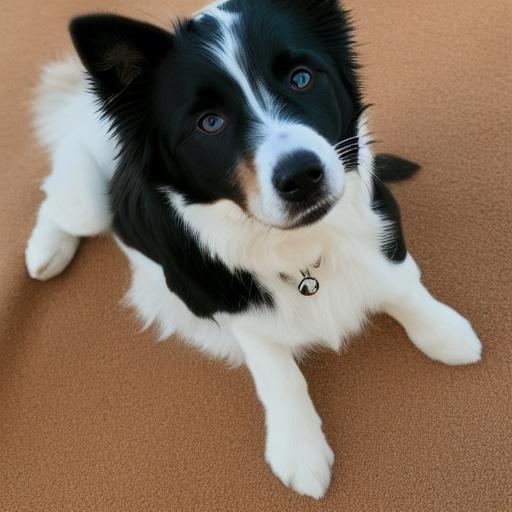} 
   \includegraphics[width=0.16\linewidth]{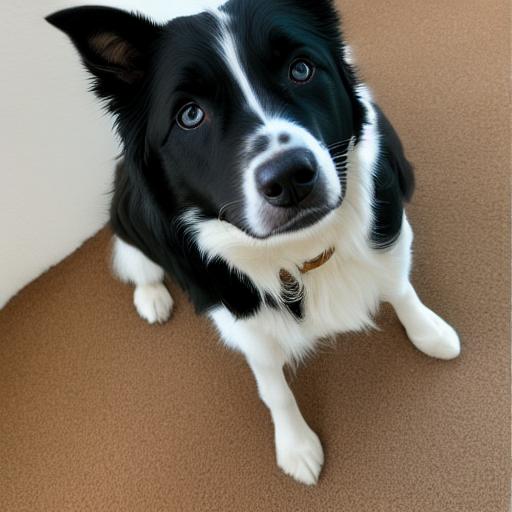}
   \includegraphics[width=0.16\linewidth]{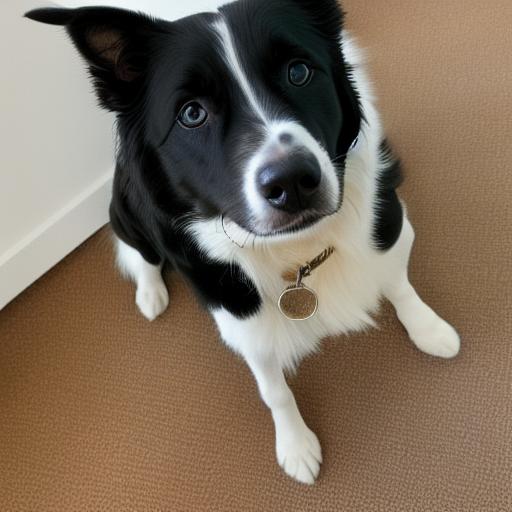} \\
       \small{~\hfill {Input} \hfill\hfill {$r=1$} \hfill\hfill {$r=6$} \hfill\hfill {$r=10$} \hfill\hfill {$r=20$} \hfill\hfill {$r=30$} \hfill ~}\\
   \caption{Image generated with the same noise and different cutoff radius $r$.  Results are based on SD1.5. }
    \vspace{-15pt}
   \label{fig:finetune-visualize}
\end{figure}

When the mask is all zero, we take a random phase for all frequencies, then this FFS noise becomes Gaussian noise and \methodname{} becomes standard diffusion. 
Figure \ref{fig:noise-visualize} visualizes FSS noise with different cutoff radius $r$. We can see that the noise becomes increasingly more structured with increasing $r$. Figure \ref{fig:finetune-visualize} shows images generated from the same input with different cutoff radius where the generated image aligns more tightly to the input with larger $r$. 

\subsection{Training Objective}
\methodname{} does not depend on model architecture or diffusion formulation. In the experiment section, we demonstrate integration both DDPM and Flow Matching, without modifying their architectures of loss functions. 

The flow matching objective learns a vector field that transports the structured noise distribution to the target image distribution. During training, given a target image $I$ and a structured noise $\hat{\epsilon}$, and a timestep $t\in[0,1]$, an intermediate image $x_t$ is obtained using a linear combination between $I$ and $\hat{\epsilon}$ following Rectified Flows \cite{liu2022flow}:
\begin{equation}
    x_t = t\ \hat{\epsilon} + (1-t)\ I. 
\end{equation}
The ground-truth velocity is 
\begin{equation}
    v_t = \frac{dx_t}{dt} = \hat{\epsilon} - I. 
\end{equation}
With this ground-truth, we can then train the model by minimizing the mean squared error between the model output and the ground-truth:
\begin{equation}
    \mathcal{L} = \mathbb{E}_{I,\hat{\epsilon},t}\| u(x_t,t;\theta)-v_t\|_2^2. 
\end{equation}
In the Fourier domain, when using full phase-preservation, the velocity becomes
\begin{equation}
    F_{v_t} = (A_{\hat{\epsilon}} - A_I) e^{j \phi_I}
\end{equation}
which has phase $\phi_I$ or $\phi_I+\pi$, preserving structural alignment. With FSS noise, phase is preserved within the cutoff radius $r$, providing controllable alignment strength.  

In DDPMs~\cite{ho2020denoising}, data $x_0$ is gradually corrupted by Gaussian noise:
\begin{equation}
q(x_t \mid x_{t-1}) = \mathcal{N}(\sqrt{1-\beta_t}\, x_{t-1}, \beta_t I). 
\end{equation}
The model learns the reverse process by predicting the added noise $\epsilon$ at each step using the loss
\begin{equation}
\mathcal{L}_{\text{DDPM}} = \mathbb{E}_{x_0,\epsilon,t}\!\left[\|\epsilon - \epsilon_\theta(x_t,t)\|_2^2\right],
\end{equation}
where $x_t = \sqrt{\bar{\alpha}_t} x_0 + \sqrt{1-\bar{\alpha}_t}\epsilon$ and $\bar{\alpha}_t = \prod_{s=1}^t (1-\beta_s)$.

In our formulation, we replace the Gaussian noise $\epsilon$ with structured noise $\hat{\epsilon}$ that preserves the input phase:
\begin{equation}
x_t = \sqrt{\bar{\alpha}_t}\, x_0 + \sqrt{1-\bar{\alpha}_t}\, \hat{\epsilon}.
\end{equation}
The training objective follows the same form,
\begin{equation}
\mathcal{L}_{\text{\methodname{}}} = \mathbb{E}_{x_0,\hat{\epsilon},t}\!\left[\|\hat{\epsilon} - \epsilon_\theta(x_t,t)\|_2^2\right],
\end{equation}
but the expectation is now over structured noise $\hat{\epsilon}$ rather than Gaussian noise $\epsilon$. 
Although $\hat{\epsilon}$ is non-Gaussian and formally violates standard DDPM assumptions, we observe high-quality, structure-aligned outputs using standard samplers after finetuning with this objective. 

\subsection{Extension to Videos}
\methodname{} extends to video by constructing phase-preserving noise frame-by-frame. For a video $\{I_1, I_2, ..., I_T\}$, we construct structured noise for each frame and concatenate along the time dimension. Since current video diffusion models produce lower-fidelity individual frames than image models, we adopt a two-stage pipeline: generating the initial frame with image-based \methodname{}, then extending temporally with first-frame-conditioned video \methodname{}. No architectural changes are required.

\section{Experiments}
\begin{figure*}[t]
  \centering
   \includegraphics[width=0.24\linewidth]{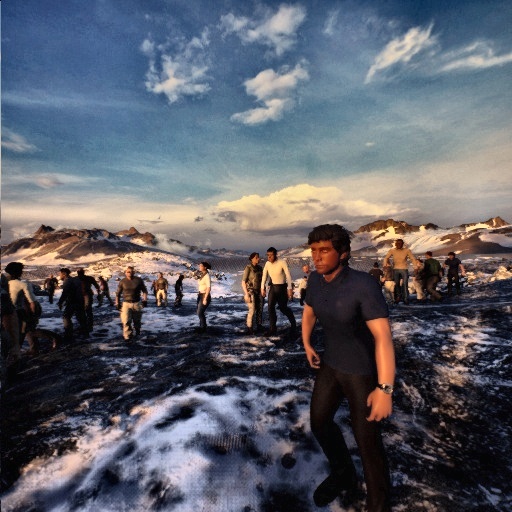} 
   \includegraphics[width=0.24\linewidth]{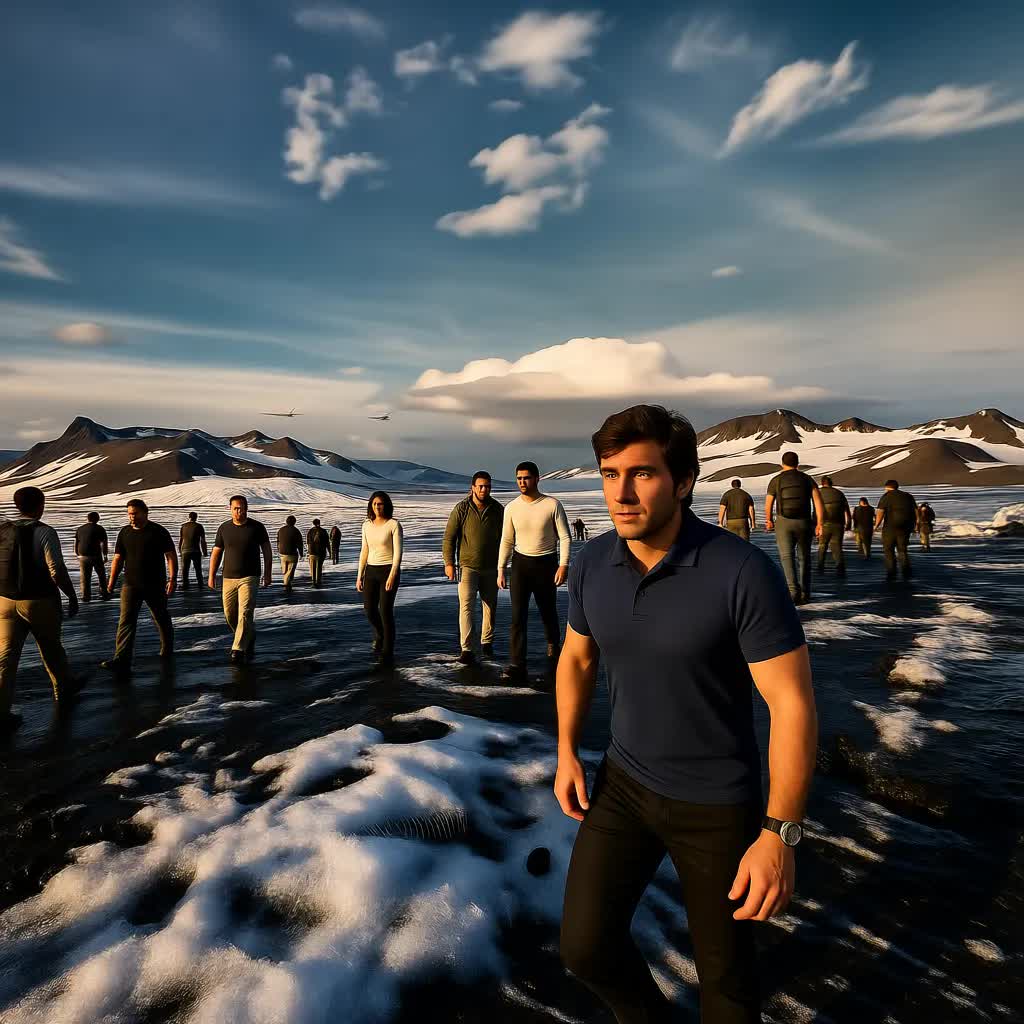} 
   \includegraphics[width=0.24\linewidth]{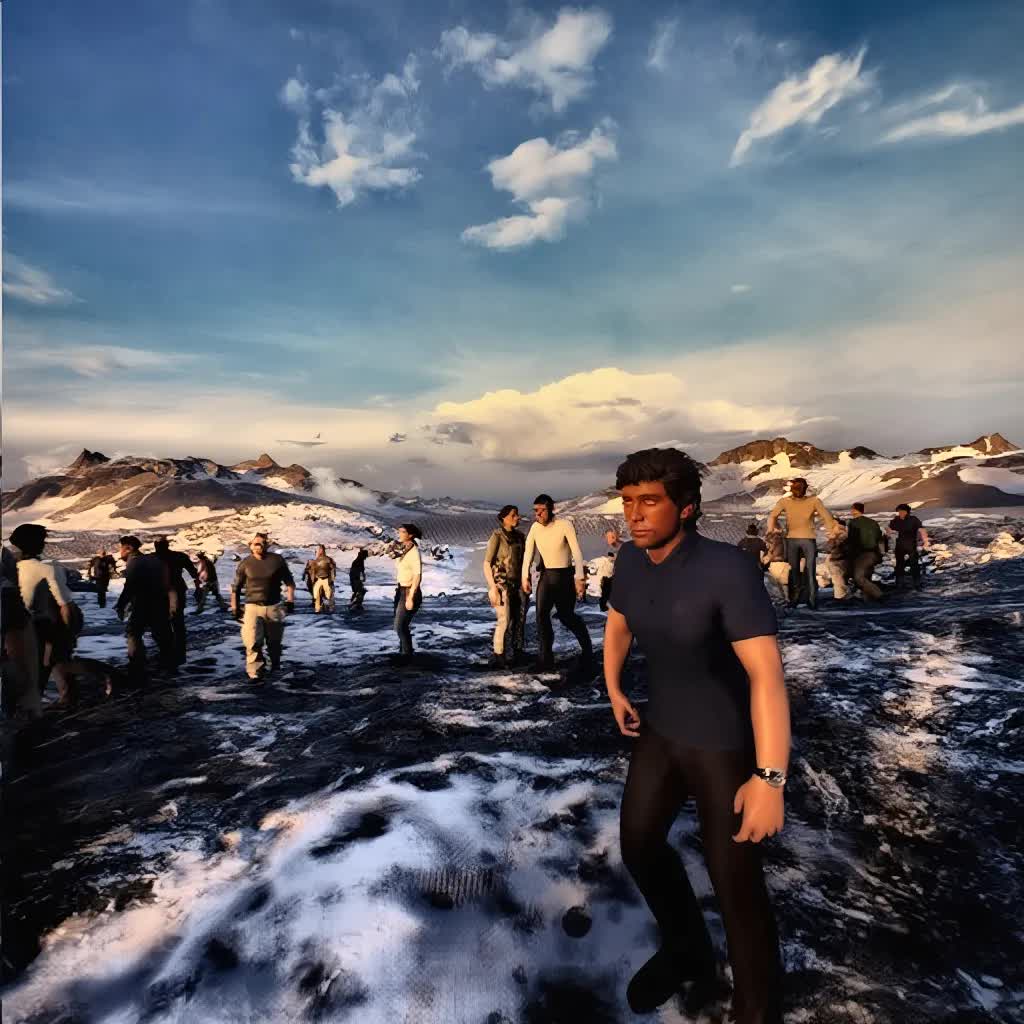} 
   \includegraphics[width=0.24\linewidth]{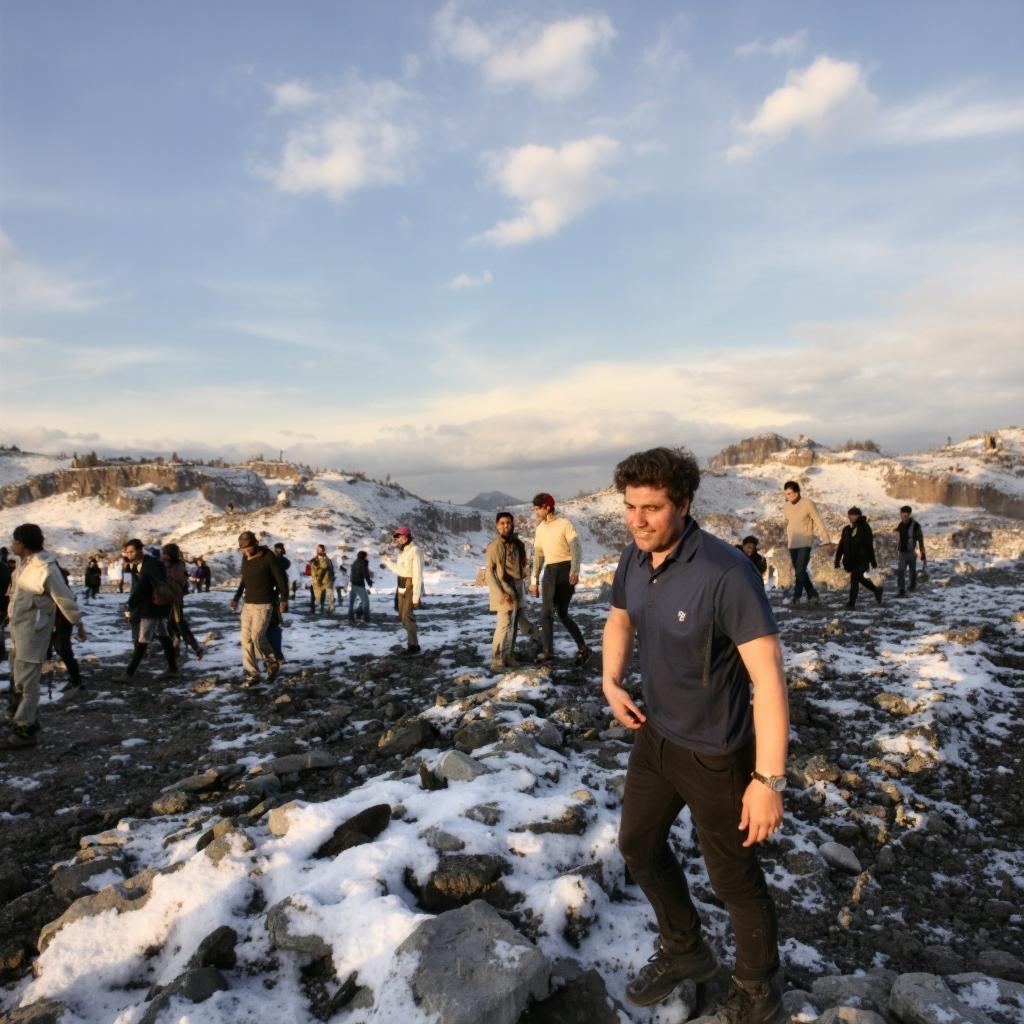} \\
   \includegraphics[width=0.24\linewidth]{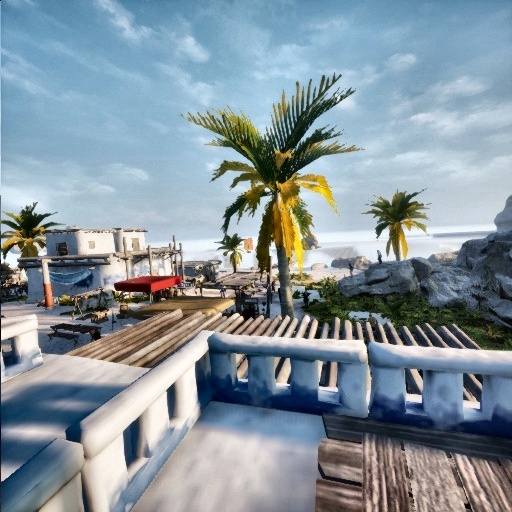} 
   \includegraphics[width=0.24\linewidth]{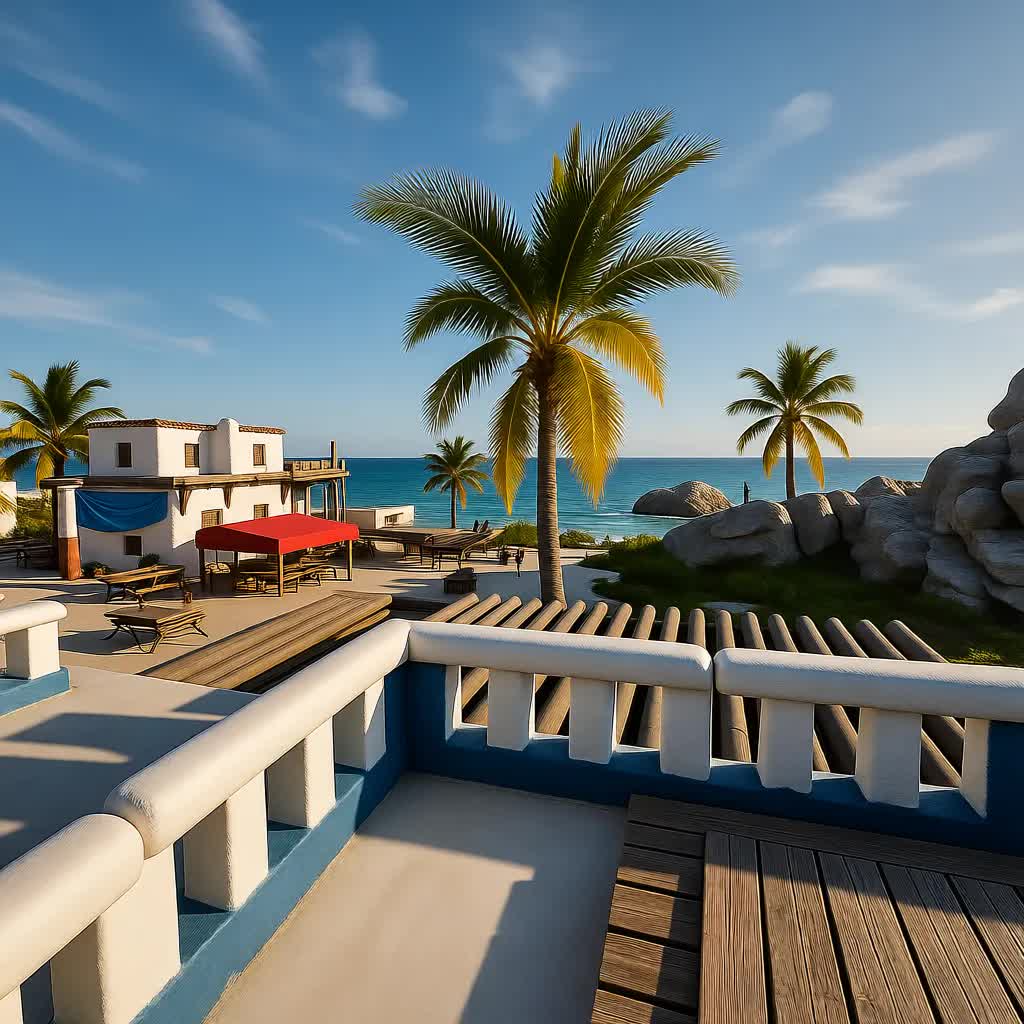} 
   \includegraphics[width=0.24\linewidth]{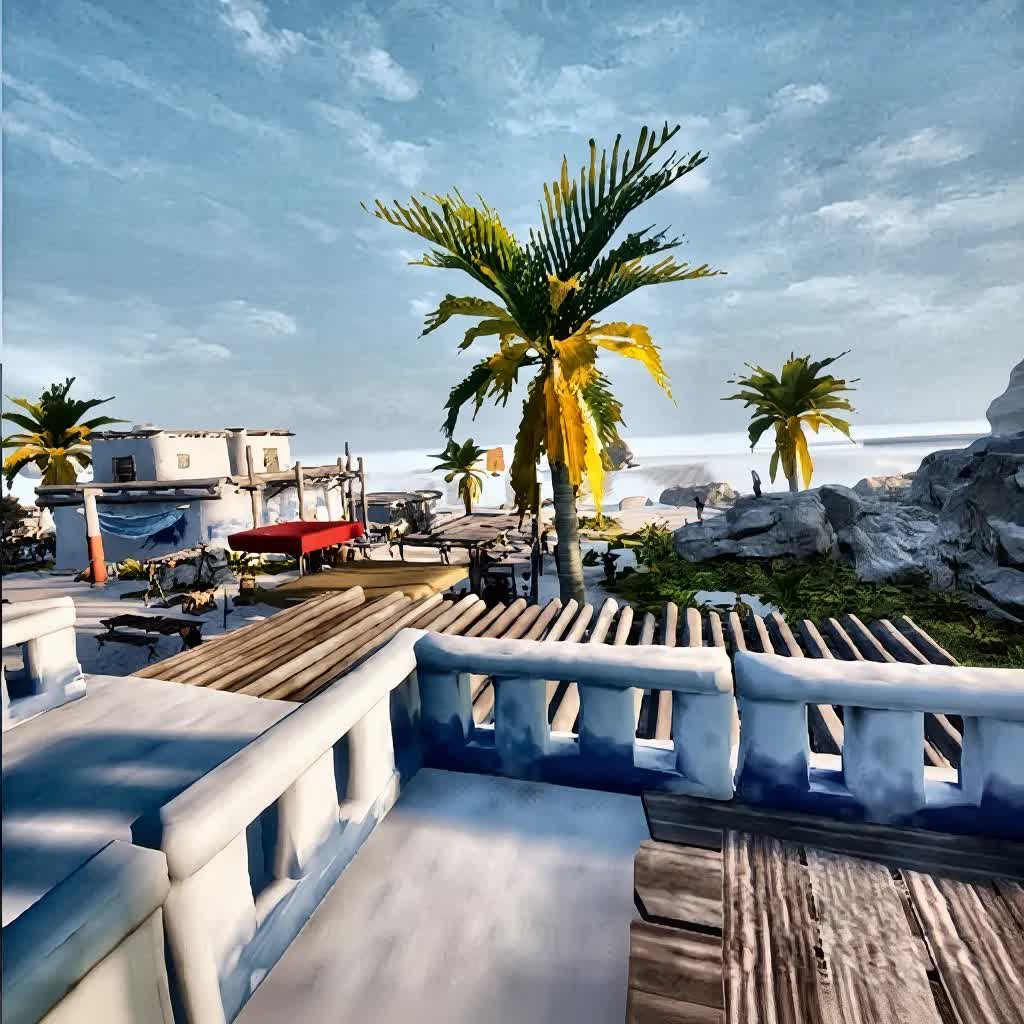} 
   \includegraphics[width=0.24\linewidth]{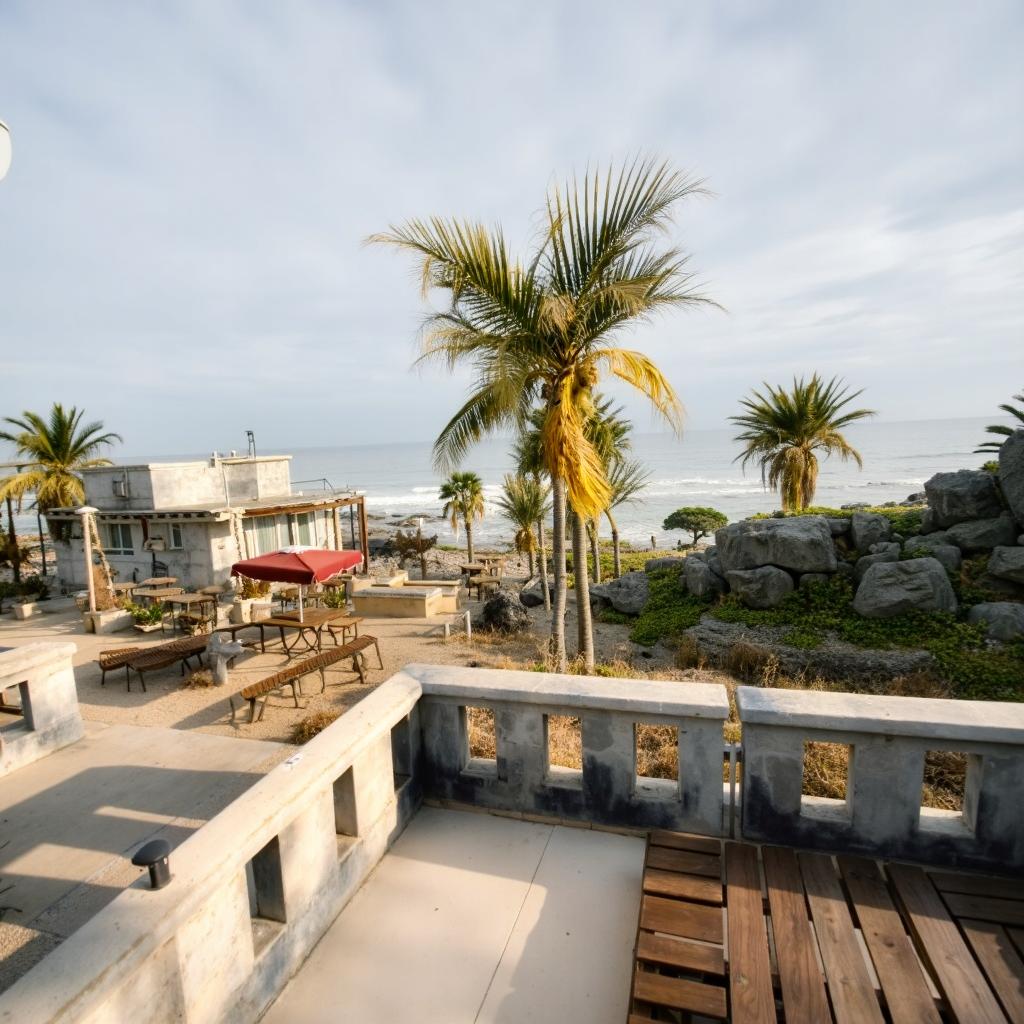}  \\
       \small{~\hfill {Input} \hfill\hfill {QWen-Edit} \hfill\hfill {FLUX.1 Kontext} \hfill\hfill {Ours} \hfill ~}\\
   \caption{Results on UnrealCV compared to FLUX-Kontext and QWenEdit.  }
    \vspace{-10pt}
   \label{fig:unrealcv}
\end{figure*}
\begin{figure*}[t]
  \centering
   ``\textbf{Pencil Sketch} of a Castle. "\\
   \includegraphics[width=0.19\linewidth]{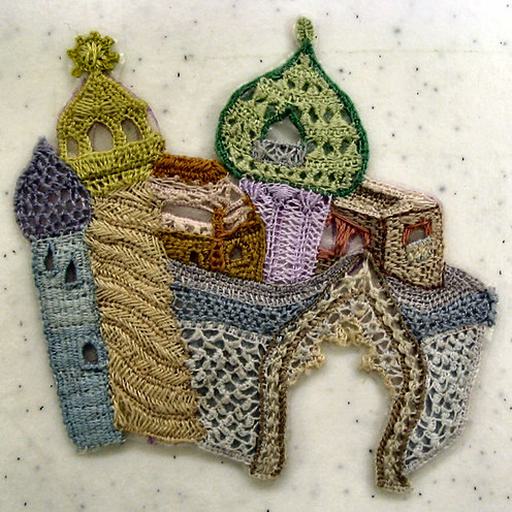}
   \includegraphics[width=0.19\linewidth]{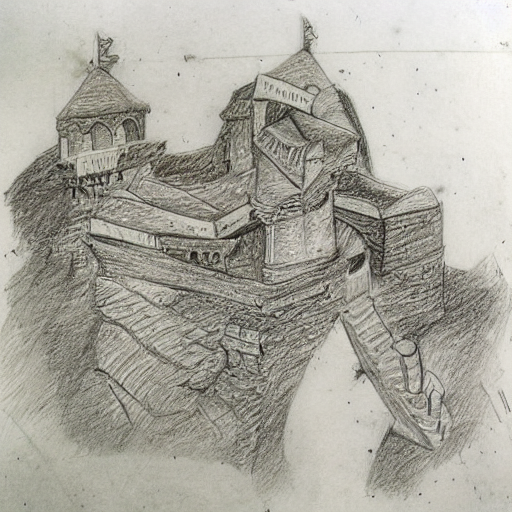}
   \includegraphics[width=0.19\linewidth]{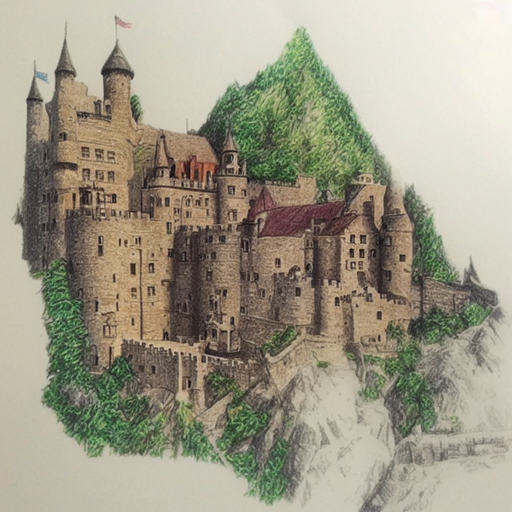}
   \includegraphics[width=0.19\linewidth]{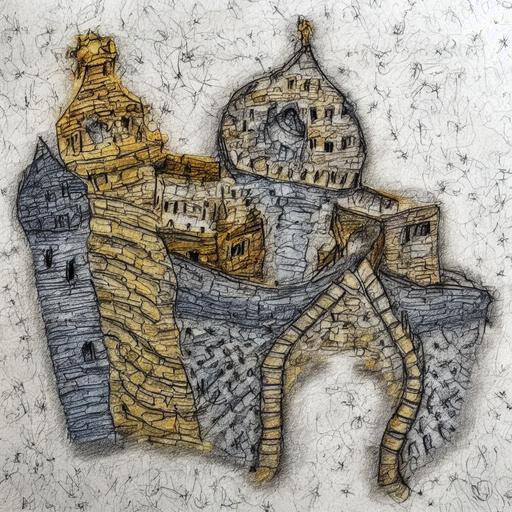}
   \includegraphics[width=0.19\linewidth]{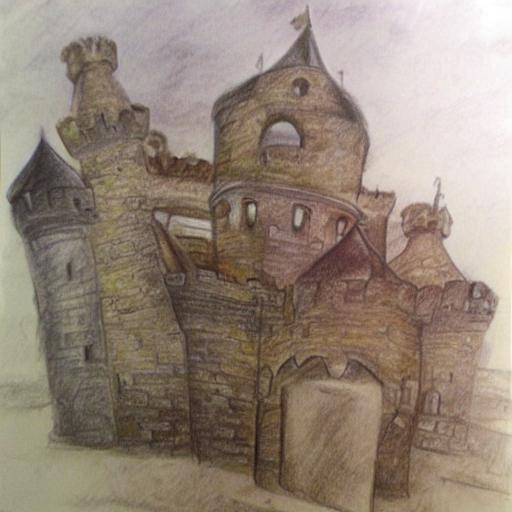}  \\
   ``\textbf{Picture} of a Husky"\\
   \includegraphics[width=0.19\linewidth]{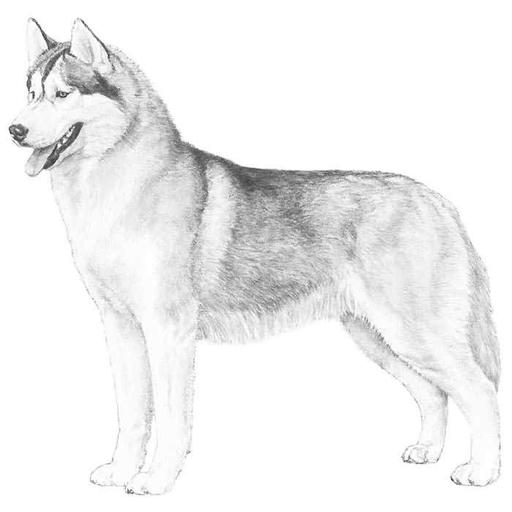}
   \includegraphics[width=0.19\linewidth]{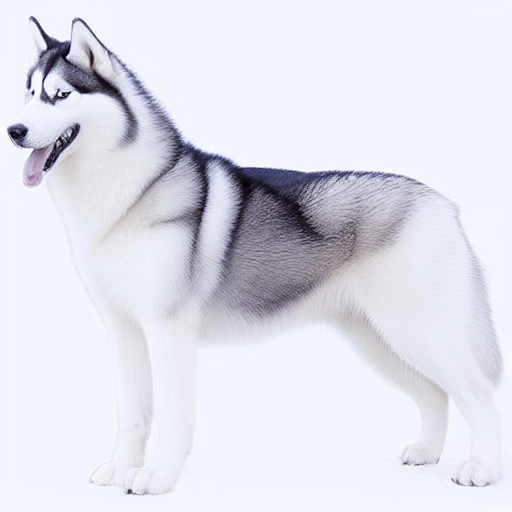}
   \includegraphics[width=0.19\linewidth]{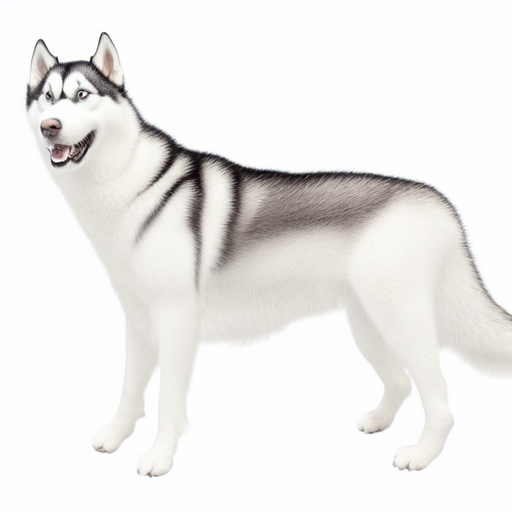}
   \includegraphics[width=0.19\linewidth]{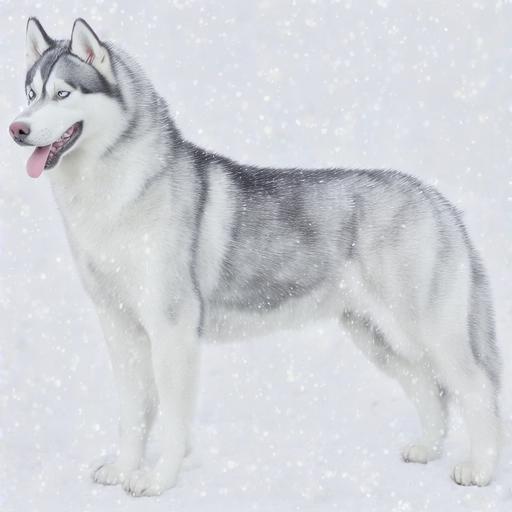}
   \includegraphics[width=0.19\linewidth]{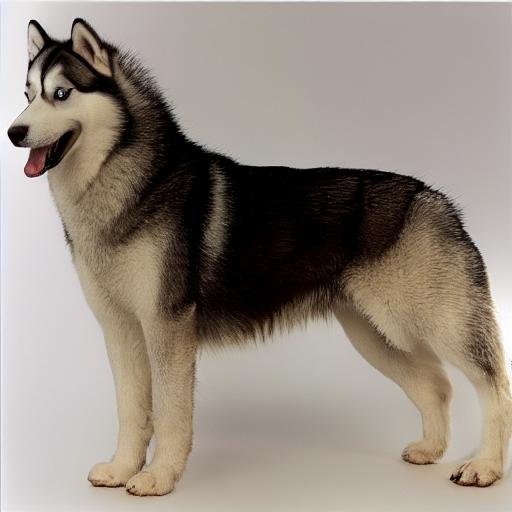}  \\
       \small{~\hfill {Input} \hfill\hfill {PNP} \hfill\hfill {SDEdit} \hfill\hfill {ControlNet} \hfill\hfill {Ours} \hfill ~}\\
   \caption{Stylized re-rendering results.  }
    \vspace{-15pt}
   \label{fig:imagenetre}
\end{figure*}
We evaluate \methodname{} across three settings: photorealistic re-rendering, stylized re-rendering, and simulation enhancement for autonomous driving, comparing against state-of-the-art methods. 
To demonstrate its broad applicability, we implement \methodname{} on three representative diffusion models: SD~1.5, FLUX-dev, and WAN~2.2 14B, which vary in size, formulation, and modality, covering both image and video generation. Please refer to the supplementary materials for additional results and code. 

During training, we use real images: construct structured noise from the image's phase, corrupt the image, and train the model to denoise. At inference, \methodname{} supports two inference modes. (1)~Structured SDEdit: add 
structured noise to the source image at some noise level $t$, then denoise. (2)~From noise: denoise from pure structured noise without any pixel contribution from the source; structure is encoded entirely in the phase. 
Mode (1) generally yields stronger structural alignment; mode (2) offers more flexibility with larger appearance changes.

\subsection{Implementation Details}

\subsubsection{Datasets}

\noindent\noindent \textbf{UnrealCV}\footnote{https://github.com/unrealcv/unrealcv} is a open-source tool that includes multiple assets. We created a diverse test set consisting of around 5,000 images across all available assets, for a total of around 200 scenes. Figure~\ref{fig:unrealcv} shows examples from this test set. This dataset covers a diverse range of scenes, including outdoor and indoor, city and natural etc, with geometry diversity while lacking photorealism. This dataset evaluates photorealistic enhancement and structure preservation. 

\noindent\noindent \textbf{ImageNetR} is a test set proposed by \cite{pnp}, including 29 images of various objects and styles. While the original dataset provides prompts, these are generic image editing prompts. Since our work primarily focuses on re-rendering, we keep the editing prompts with style hints in the original dataset and added additional style prompts, resulting in a total of 8 prompts for each image. 
This benchmark assesses stylized re-rendering and structure preservation.

\noindent\noindent \textbf{CARLA} is an open-source driving simulator~\cite{dosovitskiy2017carla}. We collect 5.5 hours of driving videos from \emph{CARLA Town 4} using the simulator's default autopilot. We then split the videos into 25 second clips and annotate a caption for each clip. For simulation enhancement, we use these captions combined with the style hint ``A photorealistic video of driving".  We evaluate the effectiveness of sim-to-real transfer by testing the CARLA-trained planner on Waymo's WOD-E2E~\cite{xu2025wode2e} validation set. 

\subsubsection{Model architecture }
We integrate \methodname{} into:
    SD 1.5 (image DDPM), 
    FLUX-dev (image flow matching), and 
    Wan2.2-14B (video flow matching). 
We either fully finetune or LoRA-finetune each model using phase-preserving noise; no architectural changes are introduced. Notably, this finetuning is highly efficient: adapting the Wan2.2-14B video model with LoRA required only a single GPU while still yielding high-quality results, further demonstrating the lightweight nature of \methodname{}. 
To evaluate structure alignment, we compute LPIPS~\cite{zhang2018unreasonable} between the input and output pairs, as well as the error between their depth map using metrics SSIM (Structural Similarity Index Measure) and ABSREL (Absolute Relative error). To evaluate text prompt alignment, we compute CLIP~\cite{radford2021learning} similarity between the generated images and the input text prompt. 

\subsubsection{Training and Inference Details}
We start from the officially released checkpoints of SD~1.5~\cite{Rombach_2022_CVPR}, FLUX-dev~\cite{flux2024}, and Wan2.2-14B~\cite{wan2025}, and finetune each model with phase-preserving noise. 
For SD~1.5, we experiment with both full finetuning and LoRA finetuning, while for FLUX-dev and Wan2.2-14B we use LoRA finetuning due to computational constraints. At inference time, for Wan2.2-14B we adopt the 4-step LoRA from LightX2V\footnote{https://huggingface.co/lightx2v/Wan2.2-Lightning} and apply it directly on top of our finetuned LoRA weights to accelerate sampling. FLUX and SD~1.5-based models are trained on PhotoConceptBucket dataset\footnote{https://huggingface.co/datasets/bghira/photo-concept-bucket}; WAN2.2-14B based models are trained on OpenSoraPlan dataset\footnote{https://huggingface.co/datasets/LanguageBind/Open-Sora-Plan-v1.1.0/tree/main}.

We LoRA finetune Wan2.2-14B for 1{,}200 iterations and FLUX-dev for 10{,}000 iterations, while SD~1.5 is fully finetuned for 140{,}000 iterations. 
Each training run takes approximately 48 hours on an NVIDIA~A100 GPU. 

For each training iteration, we sample a cutoff radius $r$ from an exponential distribution and add a constant offset $r_0$ to ensure a minimum amount of phase information is always preserved:
\begin{equation}
r = r_0 + r', \quad r' \sim \mathrm{Exp}(\lambda),
\end{equation}
where $\lambda > 0$ is the rate parameter of the exponential distribution and $r_0$ controls the minimum cutoff. 
In our experiments, we set $\lambda = 0.1$ empirically. 
We set the transition bandwidth parameter $\sigma = 2$, which controls the smoothness of the frequency mask $M(u,v)$ around the cutoff radius~$r$.

\subsection{Results}
\begin{table}[t]
\caption{Quantitative evaluation results for photorealistic re-rendering on UnrealCV.}
    \vspace{-15pt}
\label{tab:unrealcv}
\begin{center}
\setlength{\tabcolsep}{5pt}
\begin{tabular}{lcccc}
\toprule
& Prompt Align. & \multicolumn{3}{c}{Structure Alignment} \\
\cmidrule(lr){2-2} \cmidrule(lr){3-5}
Method & CLIP $\uparrow$ & LPIPS $\downarrow$ & SSIM $\uparrow$ & ABSREL $\downarrow$ \\
\midrule
ControlNet-Tile & 0.3018 & 0.5831& 0.6804& 0.9484\\
PNP & 0.2978& 0.4076&0.6993 & 0.8870\\
FBSDiff &0.3255& 0.3558& 0.6792& 0.9362\\
SDEdit & 0.2989& 0.4540&0.6754 &0.8932\\
Ours & 0.3212& 0.2397 & 0.7481& 0.5589\\
\bottomrule
\end{tabular}
\end{center}
    \vspace{-15pt}
\end{table}
\begin{table}[t]
\caption{Quantitative evaluation for stylized re-rendering. }
    \vspace{-15pt}
\label{tab:imagenetre}
\begin{center}
\setlength{\tabcolsep}{5pt}
\begin{tabular}{lcccc}
\toprule
& Prompt Align. & \multicolumn{3}{c}{Structure Alignment} \\
\cmidrule(lr){2-2} \cmidrule(lr){3-5}
Method & CLIP $\uparrow$ & LPIPS $\downarrow$ & SSIM $\uparrow$ & ABSREL $\downarrow$ \\
\midrule
ControlNet-Tile & 0.3022 & 0.5849& 0.7508 & 0.9085\\
PNP & 0.2983& 0.3029&0.7796 & 0.8816 \\
FBSDiff & 0.3091& 0.3150& 0.7562& 0.9736\\
SDEdit & 0.3011& 0.3456&0.7588  &0.9148\\
Ours &0.3017 & 0.2759 &0.7842 & 0.8087 \\
\bottomrule
\end{tabular}
\end{center}
    \vspace{-15pt}
\end{table}
\begin{table}[t]
\caption{Efficiency comparison in extra parameters and FLOPs relative to the base model as well as wall-clock time. }
\vspace{-10pt}
\centering
\small
\begin{tabular}{lccccc}
\toprule
 & ControlNet & SDEdit & PNP & FBSDiff & Ours \\
\midrule
Extra Params & +50\% & +0\% & +0\% & +0\% & +0\%\\
Extra FLOPs & +50\% & +0\% & +100\% & +1100\%& +0\%\\
Time (s/image) &28.02 &20.31 & 20.40 &133.8   &20.22 \\
\bottomrule
\end{tabular}
\label{tab:efficiency}
\vspace{-15pt}
\end{table}

\noindent \textbf{Photorealistic Re-Rendering.}
Quantitative results on UnrealCV re-rendering are summarized in Table~\ref{tab:unrealcv}, where all methods are implemented using the SD~1.5-based models for fair quantitative comparison. 
We evaluate SDEdit with Gaussian noise (SDEdit) and structured noise (Ours) at identical noise levels ($t{=}0.5$). This comparison shows that, compared to Gaussian noise, structured noise consistently improves structure preservation (near 90\% improvements on LPIPS) while maintaining text prompt alignment. 
Qualitative examples are shown in Figure~\ref{fig:unrealcv}, which compares our Flux-based model against stronger recent models such as FLUX-Kontext, and Qwen-Edit. Across both settings, all methods improve photorealism-as reflected by higher AS scores than the input images while \methodname{} achieves the highest photorealism and superior structure alignment. 
We observe that QWen-Edit produces visually high-quality results but often fails to maintain structural alignment with the input image; for example, in the first four cases, it enlarges the main subjects significantly. 
FLUX-Kontext aligns better with the input structure but provides only limited improvement in visual quality. Our method achieves both high visual fidelity and consistent structural alignment across frames.

\noindent \textbf{Stylized Re-rendering.} 
Results of stylized re-rendering are presented in Table~\ref{tab:imagenetre}, with representative examples shown in Figure~\ref{fig:imagenetre}. All models are based on SD 1.5. This task evaluates the model’s ability to alter appearance while preserving scene structure. As shown, \methodname{} produces visually coherent stylizations that maintain object boundaries and spatial consistency, while prior methods often distort geometry or introduce texture misalignment. Quantitatively, \methodname{} achieves similar prompt alignment and significantly higher structure alignment. 

\begin{wrapfigure}{r}{0.5\textwidth}
\vspace{-20pt}
   \includegraphics[width=\linewidth]{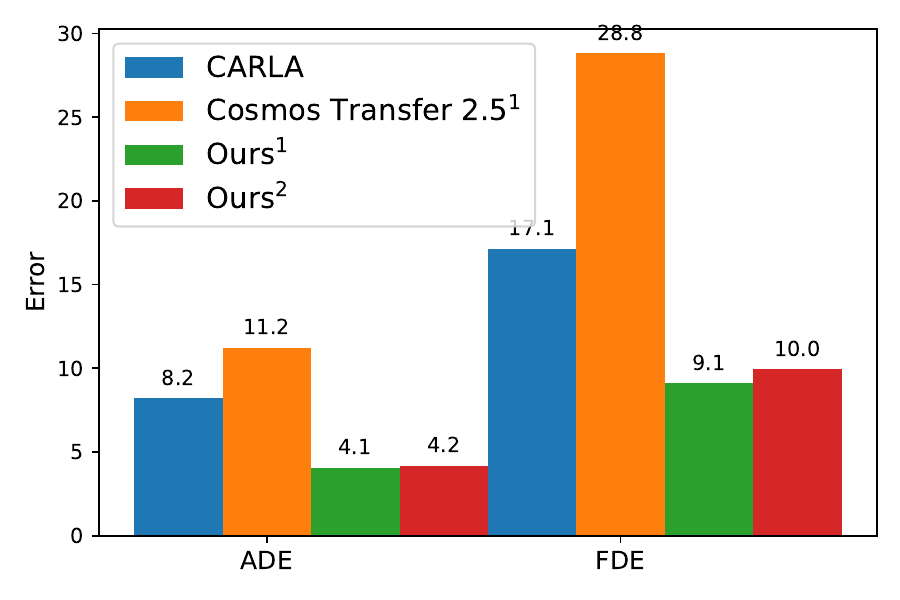}
    \caption{Planner error on Waymo validation set. Lower is better. $^1$zero-shot, $^2$finetuned on Waymo training set videos. }
\label{tab:waymo}
\vspace{-10pt}
\end{wrapfigure}
\begin{figure*}[t]
  \centering
  \begin{tabular}{@{}c@{\hspace{2pt}}c@{\hspace{2pt}}c@{\hspace{2pt}}c@{}}
   \rotatebox{90}{\parbox{1.8cm}{\centering\small Input}} &
   \includegraphics[width=0.31\linewidth]{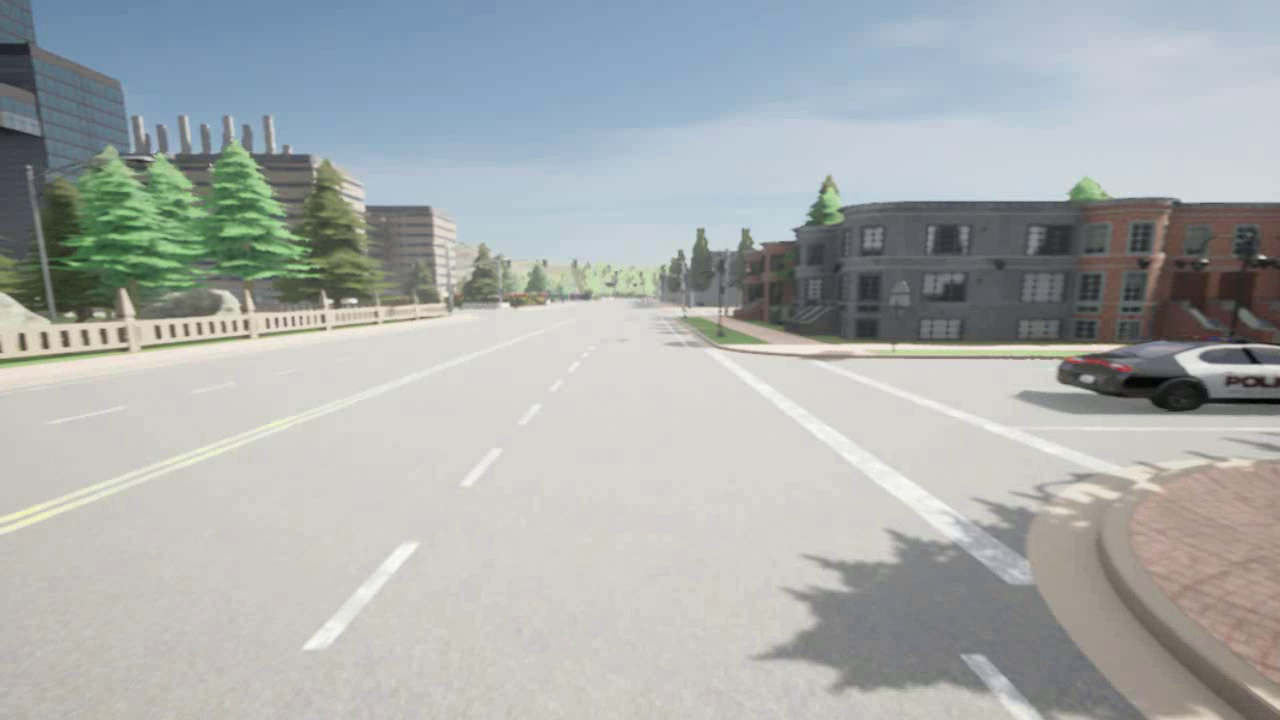} &
   \includegraphics[width=0.31\linewidth]{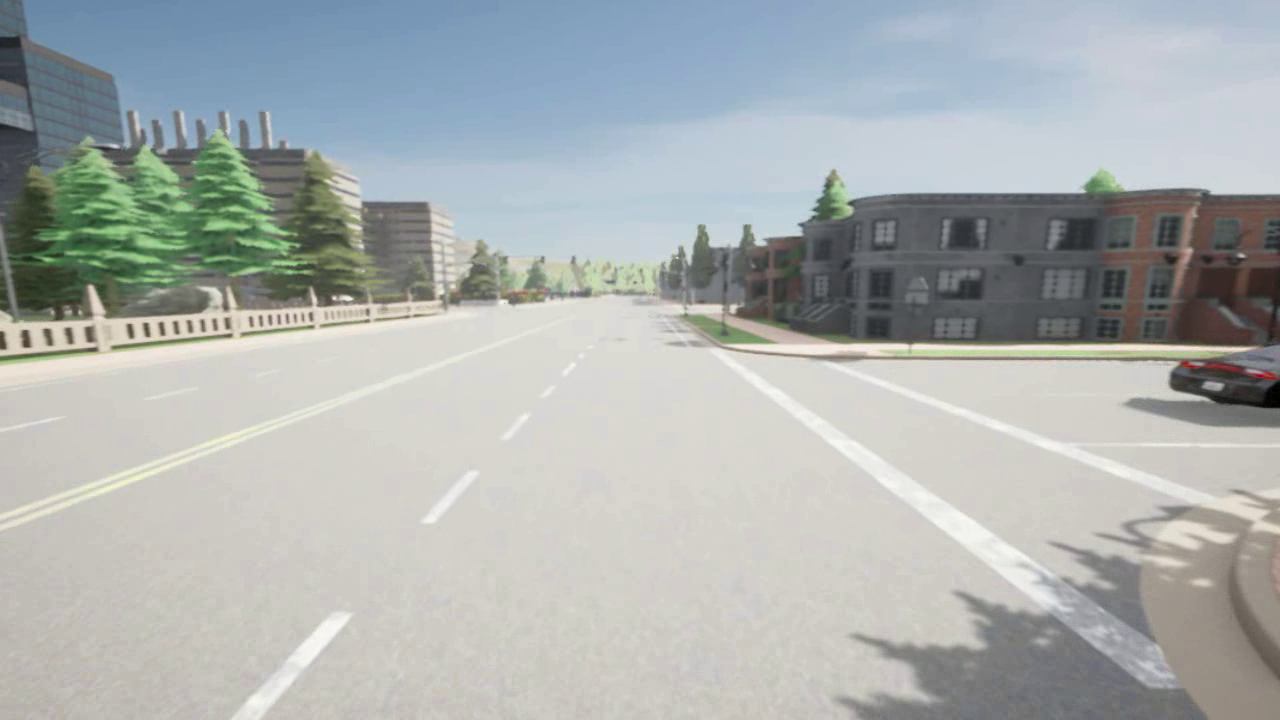} &
   \includegraphics[width=0.31\linewidth]{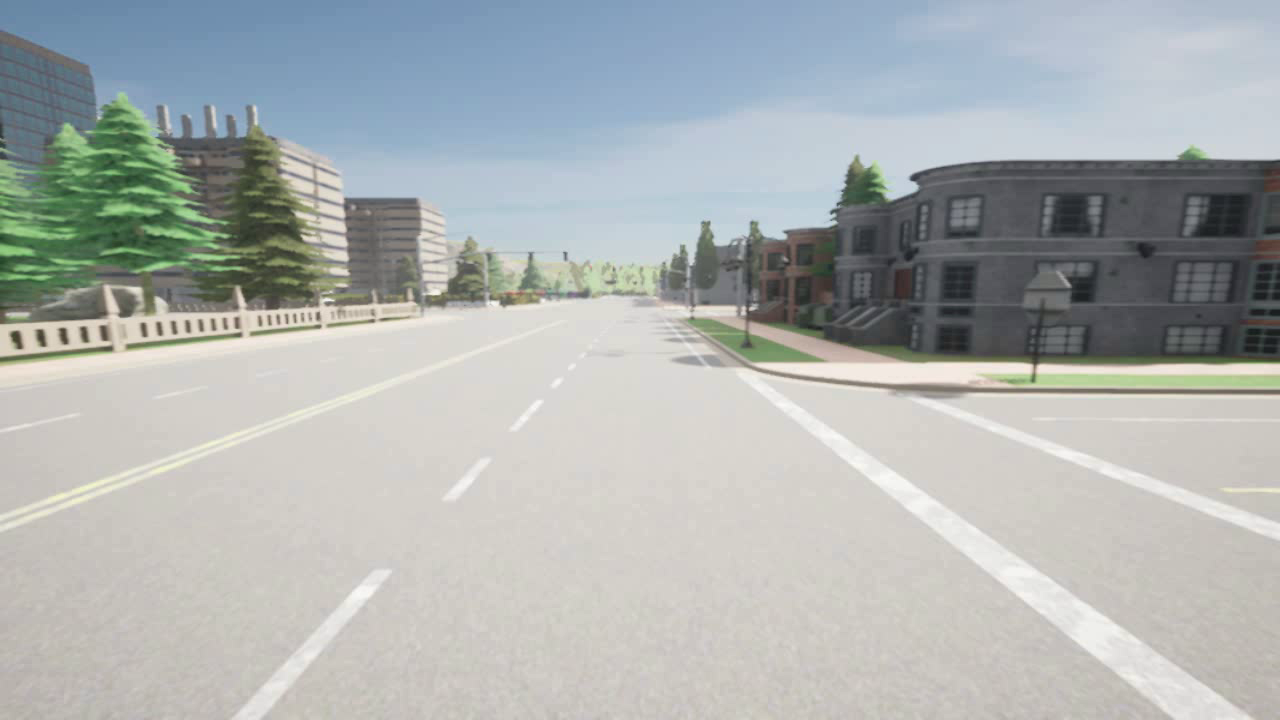} \\
   \rotatebox{90}{\parbox{1.8cm}{\centering\small Cosmos-Transfer 2.5}} &
   \includegraphics[width=0.31\linewidth]{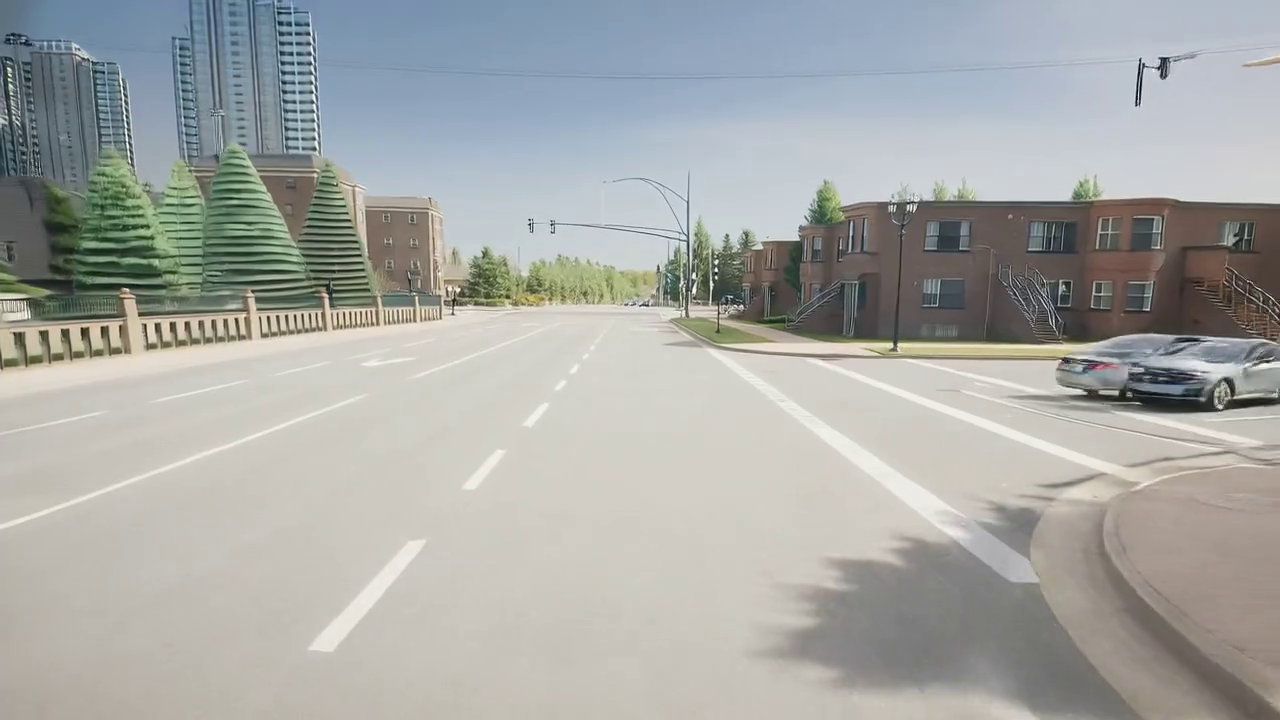} &
   \includegraphics[width=0.31\linewidth]{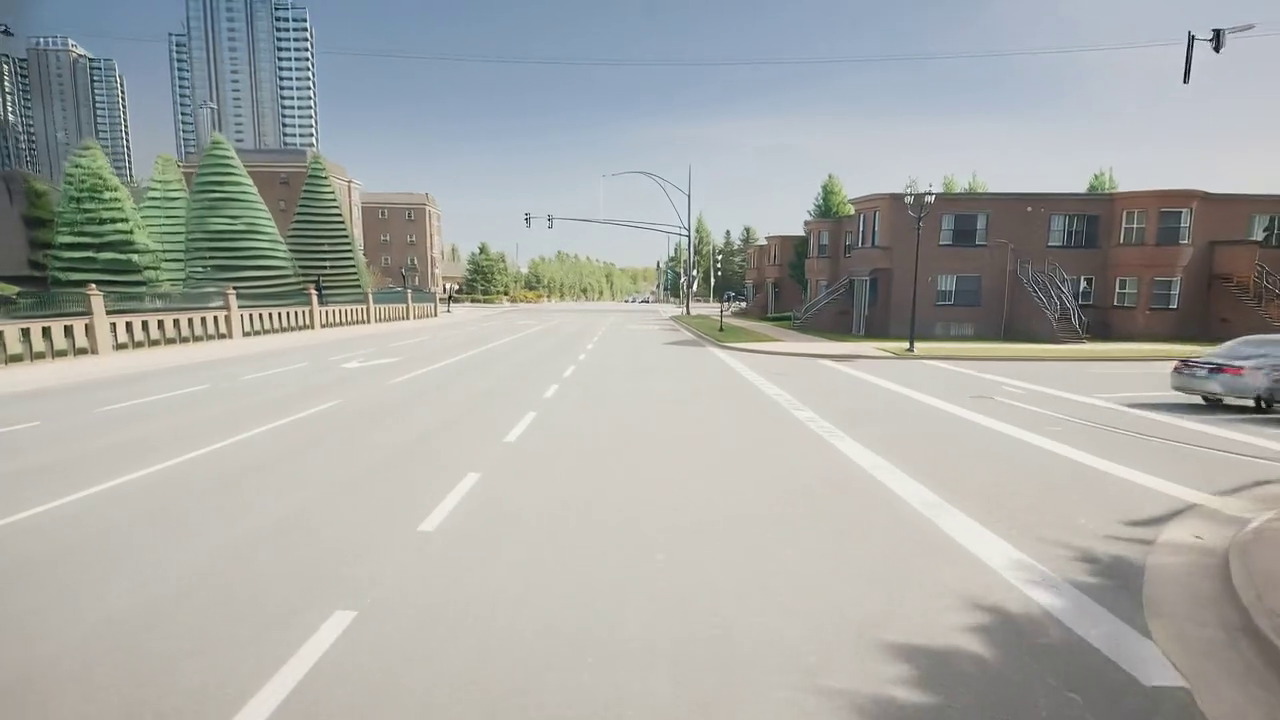} &
   \includegraphics[width=0.31\linewidth]{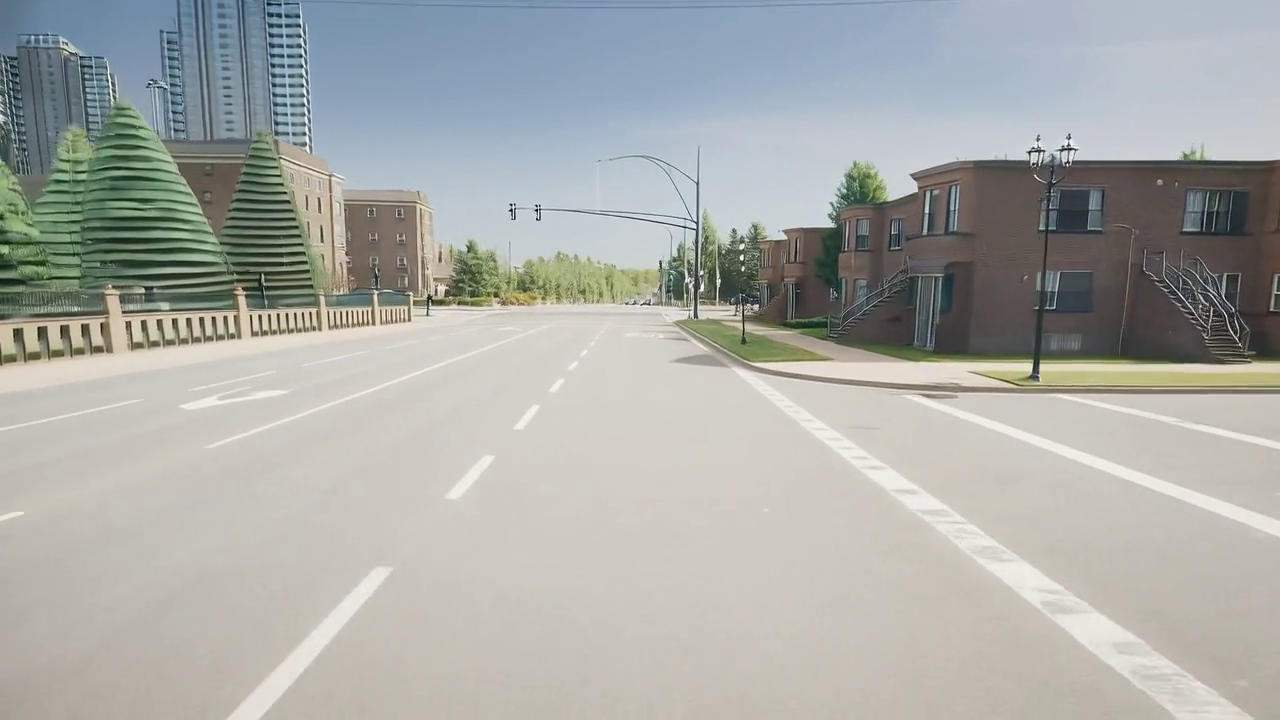} \\
   \rotatebox{90}{\parbox{1.8cm}{\centering\small Ours}} &
   \includegraphics[width=0.31\linewidth]{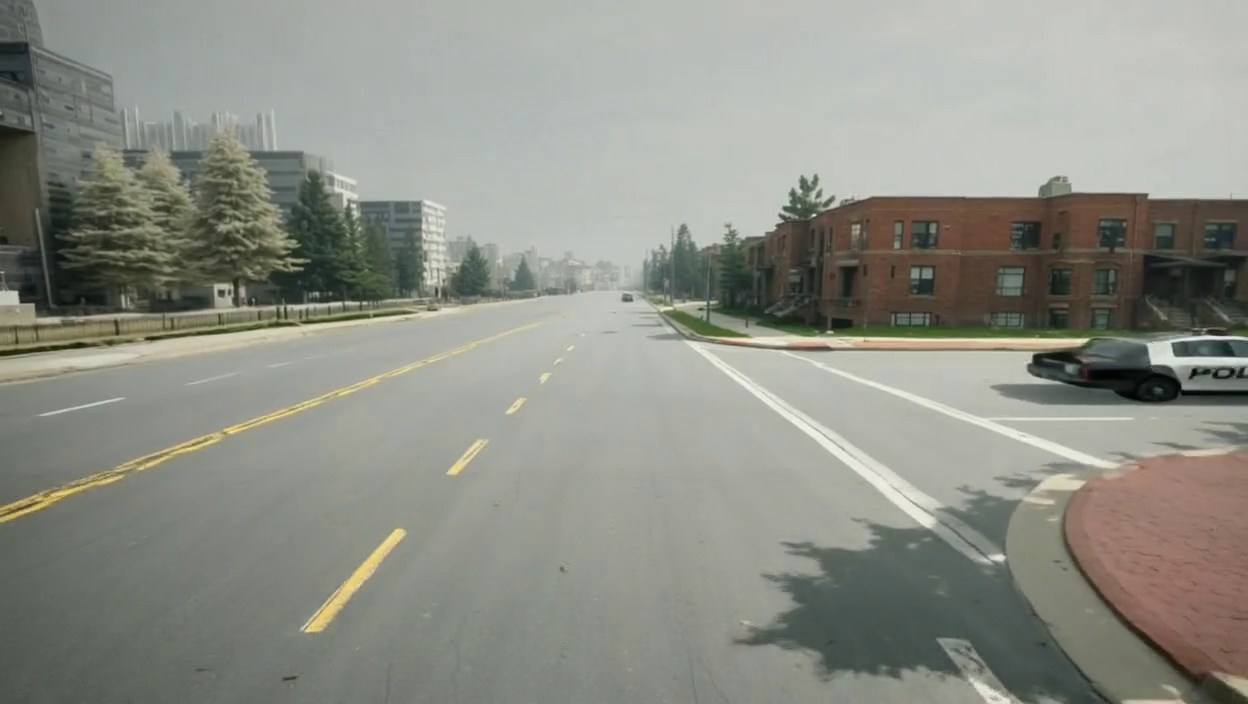} &
   \includegraphics[width=0.31\linewidth]{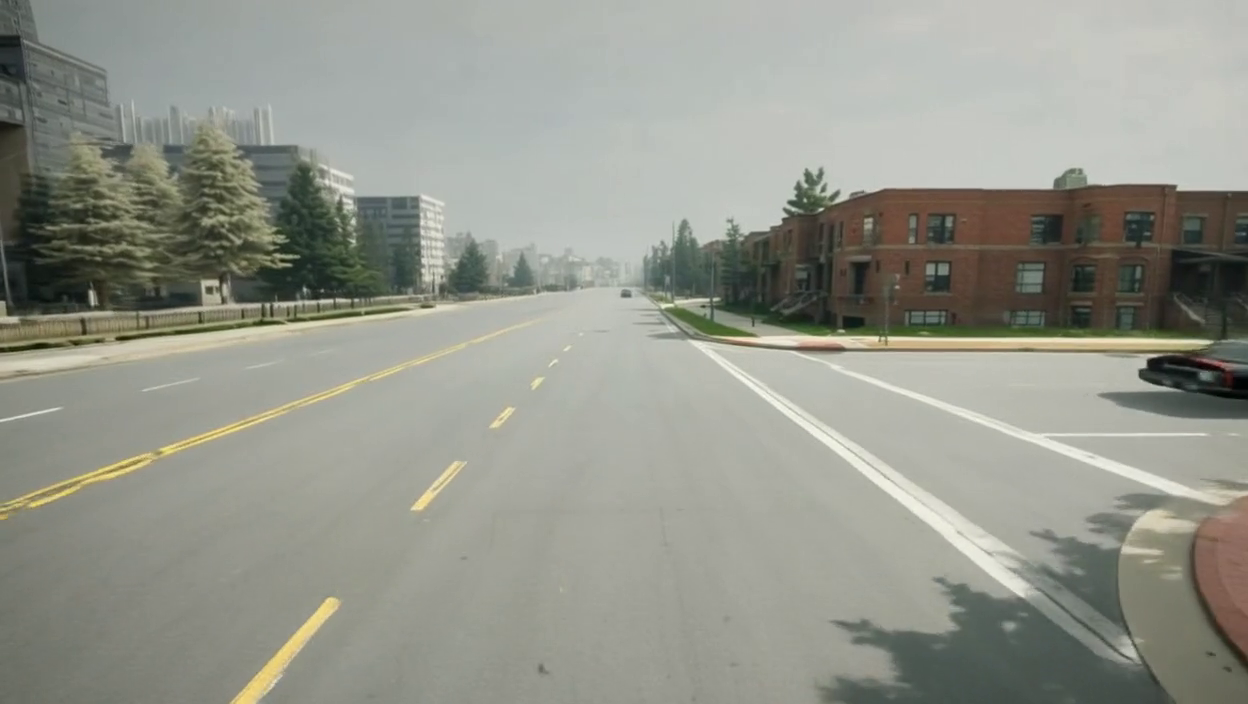} &
   \includegraphics[width=0.31\linewidth]{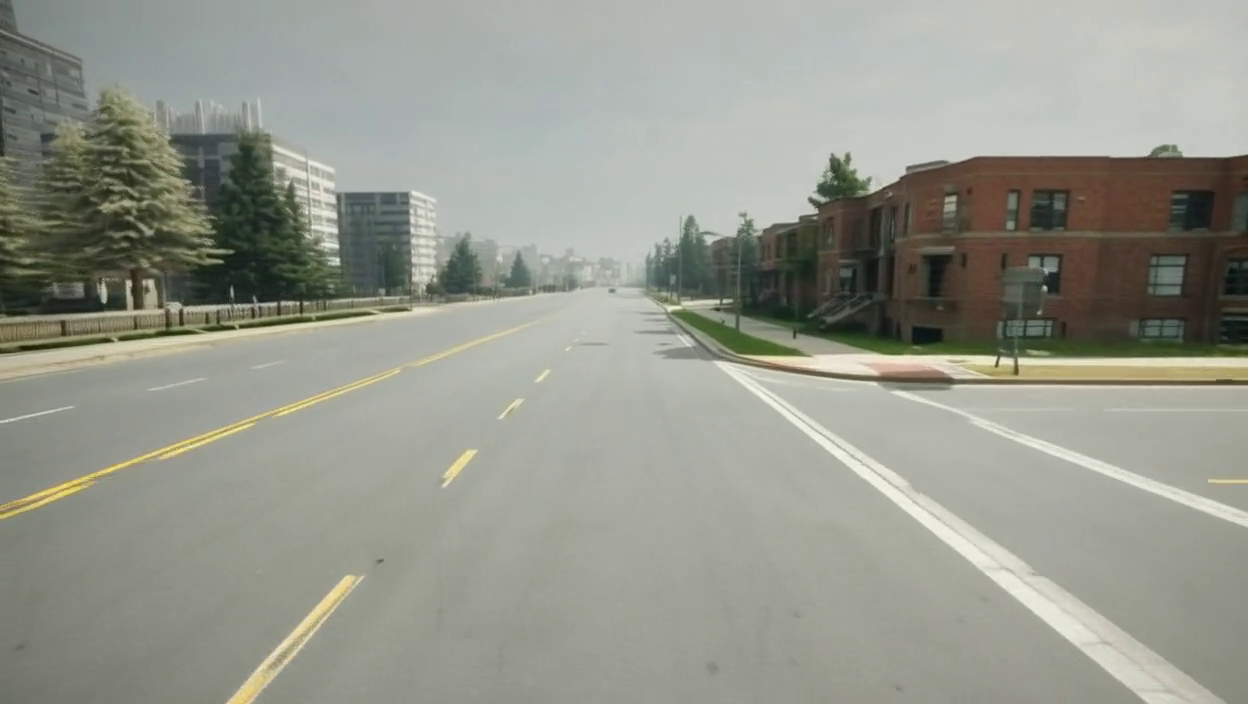} 
  \end{tabular}
   \caption{Video re-rendering results on CARLA.}
   \vspace{-10pt}
   \label{fig:waymo}
\end{figure*}

\noindent \textbf{Simulation Enhancement.}
For this experiment, we generate 5.5 hours of demonstration driving videos from CARLA using its autopilot. Then we train an end2end planner on the re-rendered CARLA videos from each method using a ResNet backbone with a GRU to take temporal input and an MLP head to output a trajectory $\in\mathcal{R}^{16\times2}$ (4s predictions at 4Hz in XY space).  As baseline, we also present the results from a purely CARLA-trained model.  
Open-loop imitation driving results are given in Figure~\ref{tab:waymo}.  \methodname{} boosts planner generalization by 50\% in zero-shot setting, demonstrating that structure-preserving appearance enhancement significantly reduces the sim-to-real gap. 
Video examples in Figure~\ref{fig:waymo} show that \methodname{} maintains road boundaries, vehicle shapes, and spatial layout consistently across frames, whereas the compared method produces distorted trees and multi-object artifacts. 

\noindent \textbf{Efficiency Comparison. }Table~\ref{tab:efficiency} compares the extra parameters and FLOPs relative to the base model, as well as the wall-clock time each method takes to process an image. ControlNet Tile adds 50\% parameters and computation overhead (UNet + ControlNet model per step. PNP is 2 $\times$ more expensive than the base model due to inversion and sampling, however achieves similar wall-clock time thanks to optimized implementation from diffuser. FBSDiff is extremely expensive (12 $\times$ base model FLOPs) due to 1000 inversion steps and 100 sampling steps. Our method does not introduce extra parameters or FLOPs. 

\begin{figure}[t!]
    \centering
    \includegraphics[width=.45\linewidth]{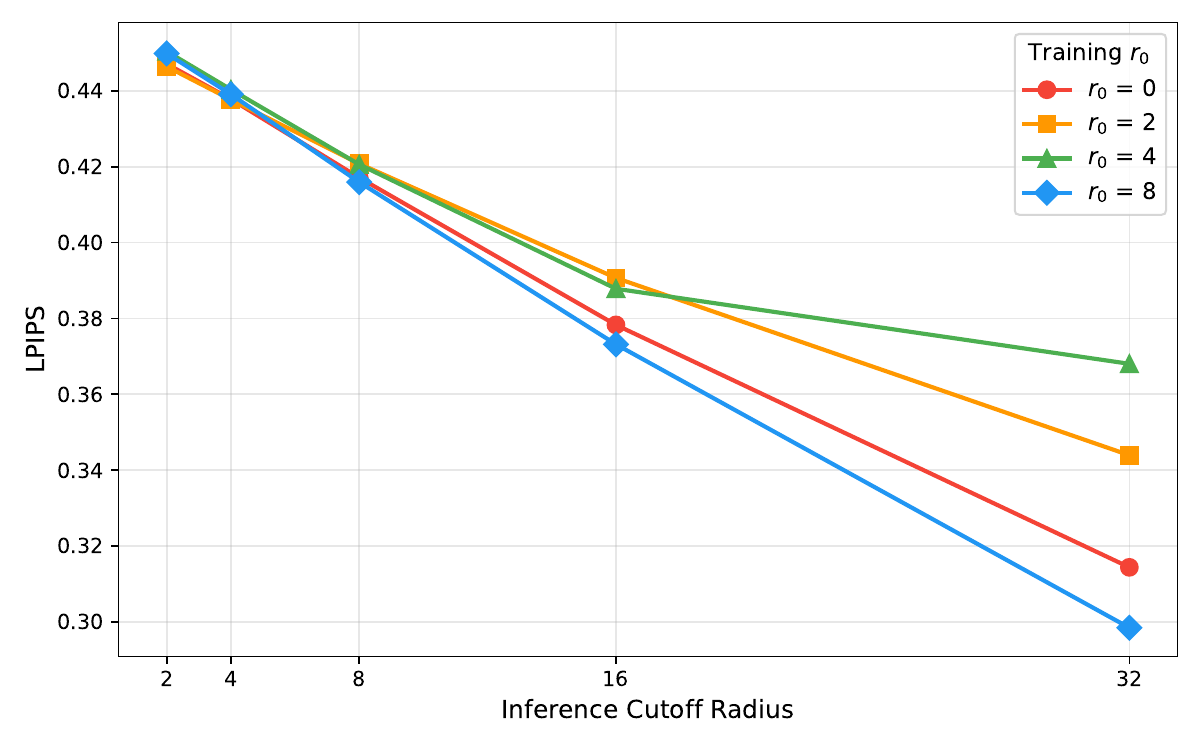}
    \includegraphics[width=.45\linewidth]{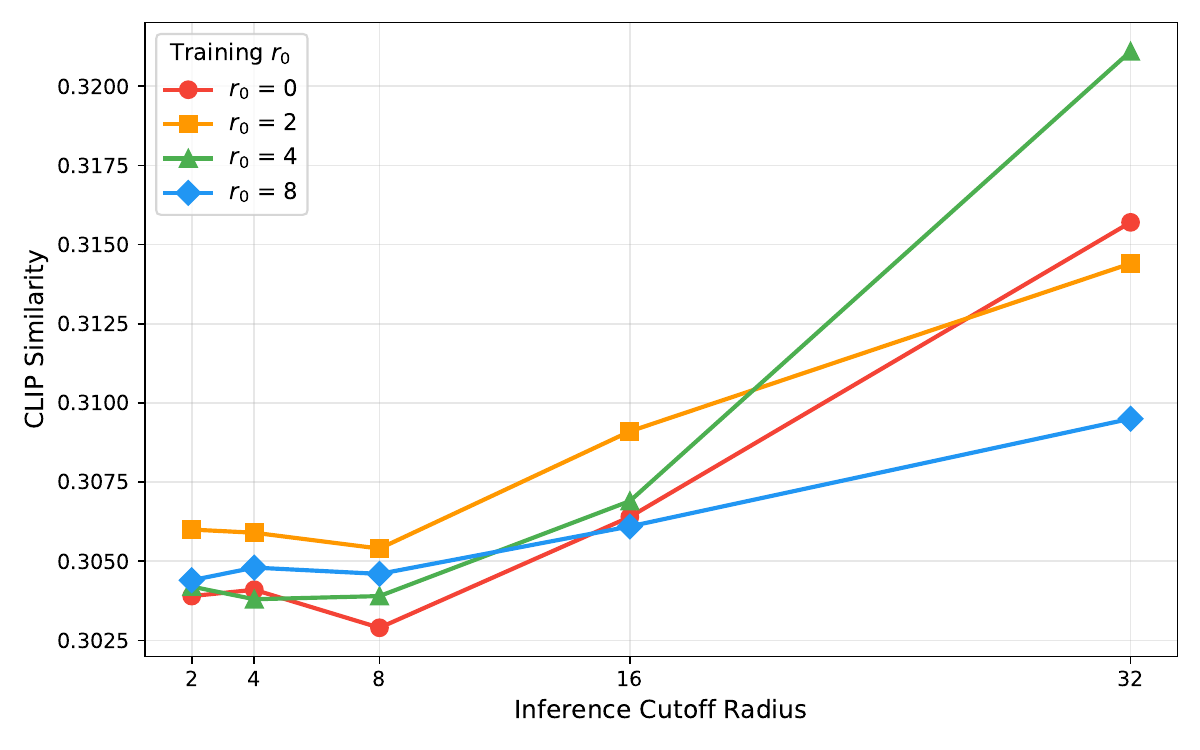}
    \caption{LPIPS and CLIP scores with different $r_0$ and $r$. }
    \label{fig:visual_ablation}
   \vspace{-10pt}
\end{figure}

\section{Ablation Studies}
\subsection{Cutoff Radius}
We ablate the choices of $r_0$, the minimal cutoff radius at training time, and the inference-time cutoff radius $r$ in this section. 
All ablation experiments are conducted using SD~1.5 with LoRA finetuning and evaluated on $1,000$ randomly selected samples from the UnrealCV test set. 
Figure~\ref{fig:visual_ablation} shows how $r_0$ and $r$ affects these metrics. 
Increasing the inference-time cutoff radius $r$ significantly improves structural alignment: LPIPS drops from $0.44-0.45$ to $0.30-0.38$, while the change in text alignment is small: CLIP scores stay around $0.30-0.32$. 
The minimal cutoff threshold $r_0$ during training influences the structural alignment with a large inference time cutoff radius $r$. It also affects performance across different inference-time radii $r$. 
A higher $r_0$ during training leads to better performance with higher $r$ during inference, while a lower $r_0$ favors scenarios with smaller inference-time $r$. 

\subsection{Phase Preservation in Latent Space.}
\begin{wrapfigure}{r}{0.5\textwidth}
\vspace{-15pt}
    \centering
    \includegraphics[width=\linewidth]{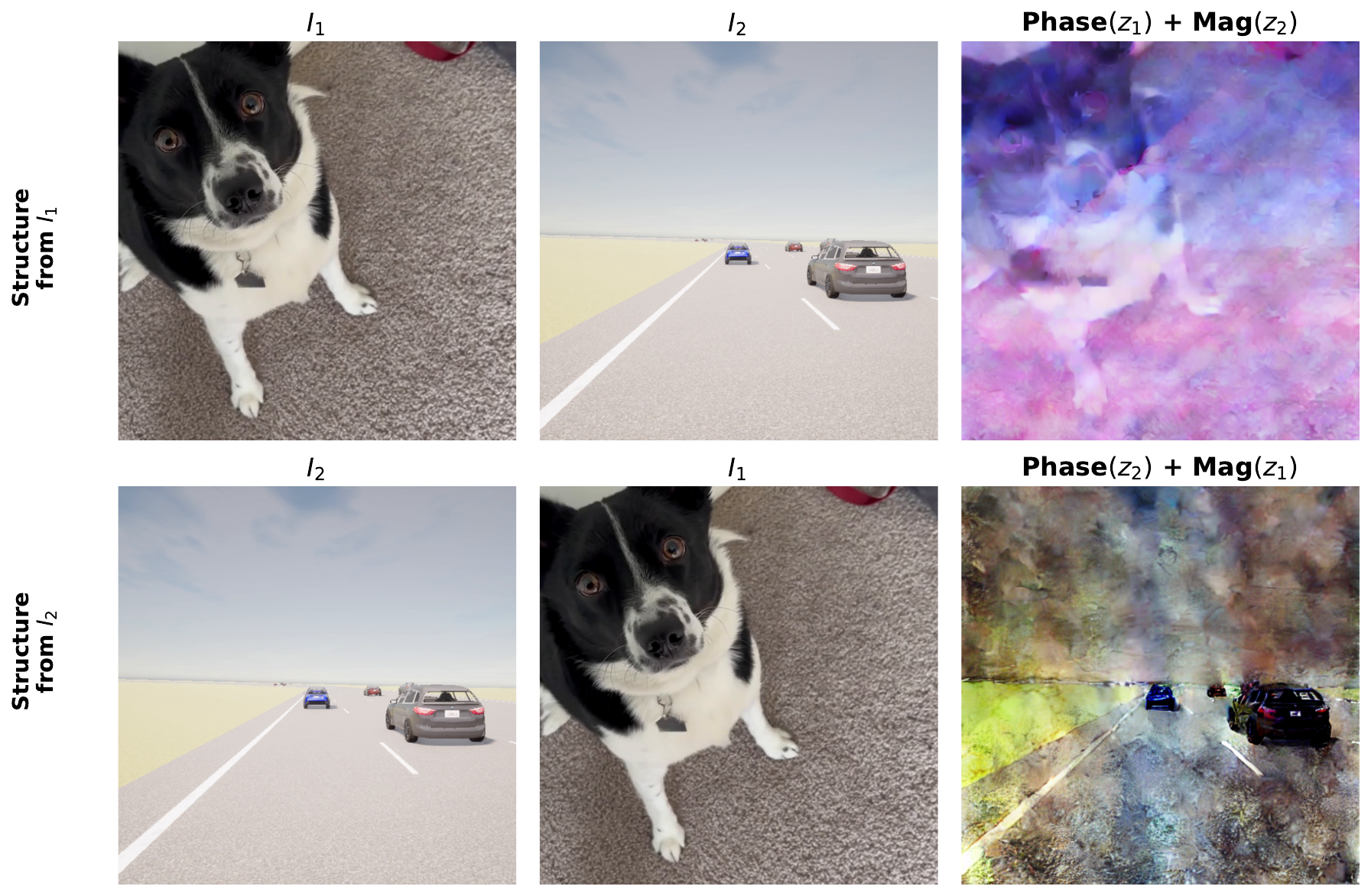}
    \caption{Phase-structure correspondence in VAE latent space. }
    \label{fig:latent_phase}
\vspace{-15pt}
\end{wrapfigure}

Classical results on phase-structure correspondence \cite{oppenheim1981importance} were established in pixel space, while modern diffusion models operate in VAE latent space. 
We apply phase preservation directly to latents. While a complete theoretical analysis of how phase-structure relationships transfer through nonlinear encoders is beyond the scope of this work, we provide empirical validation in Figure~\ref{fig:latent_phase}. 
We encode two images $I_1$ and $I_2$ into VAE latent representations $z_1$ and $z_2$, then swap their Fourier magnitude and phase components. Combining the phase from $z_1$ with the magnitude from $z_2$ produces a decoded image that preserves the structure of $I_1$ (the dog's silhouette), as shown in the top row. The bottom row shows the reverse combination which preserves the structure of $I_2$ (road, horizon, vehicles). This suggests that VAE latents, which are trained to preserve perceptually meaningful spatial structure, maintain sufficient phase-structure correspondence for our method to be effective. 

\section{Conclusion}
We introduced Phase-Preserving Diffusion (\methodname), a simple yet effective reformulation of the diffusion process that replaces Gaussian noise with structured noise, preserving image phase while randomizing the magnitude in the frequency domain. This simple change retains spatial alignment throughout sampling without modifying the architecture, altering training objectives, or introducing inference-time overhead. We also introduced Frequency-Selective Structured (FSS) noise, which provides continuous control over structural alignment rigidity through a single frequency cutoff parameter, making it broadly applicable to different applications. 

\noindent \noindent \textbf{Limitation. }\methodname{} assumes image-like inputs; modalities such as depth or normals may require a lightweight prior to produce an initial image representation. 

\noindent \noindent \textbf{Future work. } \methodname{} is orthogonal to existing conditioning or adapter methods and can be integrated with them for enhanced control. Future work includes extending \methodname{} to tasks such as deblurring, relighting, super-resolution, and general image restoration.

{\small
\bibliographystyle{ieee_fullname}
\bibliography{egbib}
}


\end{document}